\documentclass[runningheads]{llncs}
\usepackage[T1]{fontenc}
\usepackage[utf8]{inputenc}
\usepackage{amsmath,amssymb,amsfonts}
\usepackage{mathtools}
\usepackage{graphicx}
\usepackage{booktabs}
\usepackage{multirow}
\usepackage{xcolor}
\usepackage{url}
\usepackage{array}
\usepackage{makecell}
\usepackage{siunitx}
\usepackage{tikz}
\usetikzlibrary{arrows.meta,positioning,shapes.geometric,calc}
\usepackage{float}
\usepackage[hidelinks]{hyperref}

\definecolor{forgetred}{RGB}{214,96,77}
\definecolor{oraclegreen}{RGB}{26,122,60}
\definecolor{ruleraccent}{RGB}{74,111,165}
\definecolor{rulermuted}{RGB}{120,120,128}

\newcommand{\ff}{\texttt{ff}}
\newcommand{\Mone}{M_{1}}
\newcommand{\Mtwo}{M_{2}}
\newcommand{\Mthree}{M_{3}}
\newcommand{\Mfour}{M_{4}}
\newcommand{\rrb}{r_{\mathrm{rb}}}
\newcommand{\Df}{\mathcal{D}_{\mathrm{f}}}
\newcommand{\Dr}{\mathcal{D}_{\mathrm{r}}}
\newcommand{\reals}{\mathbb{R}}
\newcommand{\modelU}{\theta^{\mathrm{u}}}
\newcommand{\modelO}{\theta^{\mathrm{o}}}
\newcommand{\modelR}{\theta^{\mathrm{r}}}
\newcommand{\similarity}{\mathrm{sim}}
\newcommand{\simr}{s_{\mathrm{r}}}
\newcommand{\simf}{s_{\mathrm{f}}}

\begin{document}
\title{RULER: Representation-Level Verification of Machine Unlearning}
\titlerunning{RULER: Representation-Level Verification}
\author{Georgina Cosma\inst{1} \and Axel Finke\inst{2}}
\authorrunning{G.\ Cosma and A.\ Finke}
\institute{Department of Computer Science, Loughborough University, UK\\
\email{g.cosma@lboro.ac.uk}
\and
School of Mathematics, Statistics and Physics, Newcastle University, UK\\
\email{axel.finke@newcastle.ac.uk}}
\maketitle

\begin{tikzpicture}[remember picture, overlay]
  \node[rotate=90, anchor=center, text=rulermuted, font=\large, align=center]
    at ([xshift=-1.2cm]current page.east)
    {Accepted at the 2nd Workshop on Machine Unlearning and Privacy Preservation
     (WIPE-OUT 2026),\\ ECML PKDD 2026, Naples, Italy. To appear in Springer CCIS.};
\end{tikzpicture}

\begin{abstract}
Machine unlearning aims to remove the influence of specific training records from a deployed model without retraining from scratch. Current protocols verify this at the output level through membership inference, retain accuracy, and forget-set accuracy, but a model can satisfy all three whilst still encoding forgotten records in its intermediate representations. We introduce RULER, a set of representation-level verification metrics. The oracle-comparative metric $\Mtwo$ measures whether forget-set records occupy the same representational position as in a model retrained without them. The oracle-free metric $\Mfour$ detects residuals from the unlearned model's internal similarity structure alone, without retraining. Four approximate unlearning methods all pass output-level evaluation, yet under a linear mixed-effects model $\Mtwo$ detects significant residuals in 10 of 12 conditions ($p<0.05$), with all four methods significant at the highest forget fraction. A fifth method, Bad Teacher, shows the same residuals despite a different forgetting mechanism. $\Mfour$ acts as a pre-unlearning diagnostic across tabular, image, clinical text, and face-identity settings: it detects identity-level memorisation in models trained on face images, where no tested method returns the metric to its null.
\end{abstract}

\section{Introduction}
\label{sec:introduction}
\begin{figure}[t]
  \centering
  \includegraphics[width=\linewidth]{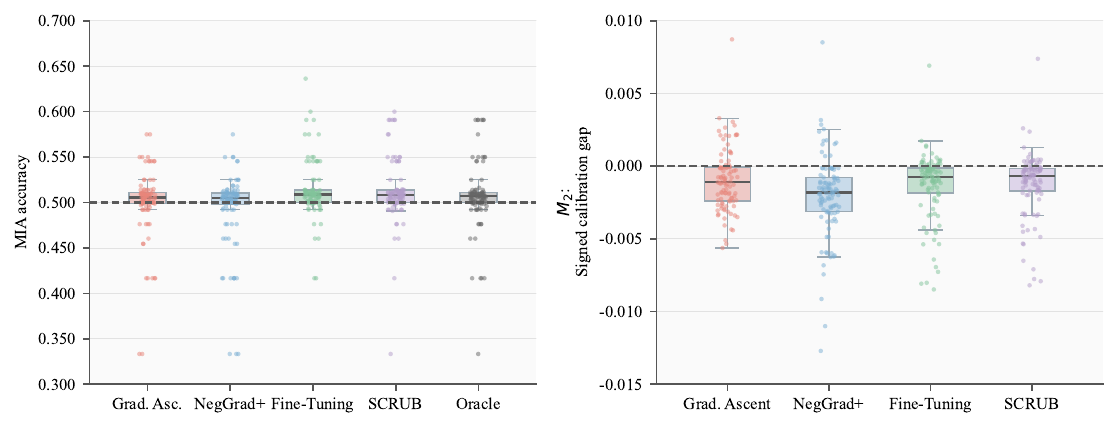}
  \caption{Discordance at $\ff=\SI{5}{\percent}$ ($N=100$ per method; 10 datasets $\times$ 10 seeds). (a)~MIA accuracy falls within the $\pm 0.05$ pass window for all methods and oracles. (b)~The signed calibration gap $\Mtwo$ (Section~\ref{sec:framework}) is negative for every method. Dashed lines: null values ($0.50$ for MIA; $0$ for $\Mtwo$).}
  \label{fig:fig0}
\end{figure}
Machine unlearning removes the influence of a specified subset of training records (the \textit{forget set}) from a trained model without retraining from scratch, whilst preserving the model's behaviour on the remaining records (the \textit{retain set})~\cite{bourtoule2021machine,cao2015towards}. Approximate methods such as gradient ascent, fine-tuning, and distillation achieve this at a fraction of the computational cost of full retraining~\cite{kurmanji2023towards}. Determining whether unlearning has succeeded requires an evaluation protocol; most existing protocols rely on three output-level criteria: (a)~membership inference attack (MIA) accuracy close to the 0.50 chance level~\cite{shokri2017membership}; (b)~preserved retain-set accuracy; (c)~forget-set accuracy matching a retrain oracle (a model trained from scratch on $\Dr$ alone)~\cite{kurmanji2023towards}. These criteria assess predictions, not the internal representations that produce them. A model can satisfy all three whilst still encoding forget-set information in intermediate layers in ways measurably distinct from a model that was never trained on those records~\cite{cf-k}. We report this discordance across four approximate methods — Gradient Ascent~\cite{golatkar2020eternal}, NegGrad$+$~\cite{kurmanji2023towards}, Fine-Tuning, and SCRUB~\cite{kurmanji2023towards} — corroborated by a fifth, Bad Teacher~\cite{chundawat2023can}: all pass output-level evaluation yet leave statistically significant traces in their internal representations (Fig.~\ref{fig:fig0}). Our contributions are as follows.

\begin{enumerate}
  \item[-] We identify a discordance between output-level and representation-level evaluation in four approximate unlearning methods.
  \item[-] We introduce RULER, a set of representation-level verification metrics: an oracle-comparative metric $\Mtwo$ that measures deviation from a retrained reference model, and an oracle-free metric $\Mfour$ that detects residuals from the unlearned model alone.
  \item[-] We show that residuals are small but directionally consistent across datasets, forget fractions, and training seeds; $\Mtwo$ remains significant under a linear mixed-effects model accounting for within-dataset correlation.
  \item[-] We demonstrate that $\Mfour$ serves as both a \textit{pre-unlearning diagnostic} and a \textit{post-unlearning check} that flags incomplete erasure or over-displacement, across tabular, image, clinical text, and face-identity settings.
\end{enumerate}

\section{Related Work}
\label{sec:related}

\paragraph{Approximate unlearning and its evaluation.}
Approximate methods modify a trained model without full retraining: gradient ascent on the forget set and fine-tuning on the retain set serve as baselines~\cite{golatkar2020eternal}, NegGrad$+$~\cite{kurmanji2023towards} combines them, and SCRUB~\cite{kurmanji2023towards} uses teacher-student distillation. Evaluation typically relies on output-level criteria relative to a retrain oracle: MIA accuracy, retain accuracy, and forget accuracy~\cite{golatkar2020eternal,kurmanji2023towards,shokri2017membership}. Hayes et al.~\cite{hayes2024inexact} showed that unlearning MIAs can overestimate privacy protection; Goel et al.~\cite{cf-k} demonstrated that indistinguishability protocols are incomplete; Thudi et al.~\cite{thudi2022auditable} argued that parameter-space comparisons cannot certify unlearning. Wang et al.~\cite{Wang2025EMUModelDifference} propose evaluating via model differences using influence functions. Chundawat et al.~\cite{chundawat2023can} introduce the Zero Retrain Forgetting (ZRF) metric, which measures Jensen--Shannon divergence between the unlearned model's forget-set outputs and a randomly initialised teacher, alongside the Bad Teacher framework used in our evaluation.

\paragraph{Representation-level analysis.}
Golatkar et al.~\cite{golatkar2020eternal} introduced weight scrubbing, observing that output metrics may not reflect whether information persists in the network; feature spaces are known to encode structured information not visible from outputs~\cite{kornblith2019similarity}. Existing metrics operate at the parameter or output level: activation distance~\cite{chundawat2023can,golatkar2020forgetting} compares softmax outputs (despite its name, not an intermediate-layer measure); completeness~\cite{cao2015towards} measures predicted-class agreement; MIA~\cite{shokri2017membership} assesses membership leakage. None examine how a model internally organises information about individual records, or how forget records are treated relative to retain records. RULER addresses this gap with per-record penultimate-layer metrics calibrated against partition-aware nulls ($\Mtwo$, $\Mfour$).

\paragraph{Tabular data.} Much of the unlearning literature focuses on image classification~\cite{golatkar2020eternal,kurmanji2023towards}; tabular data instead poses heterogeneous features, class imbalance, and small forget sets that limit statistical power, so we use it as the primary setting and extend to image, clinical text, and face-identity domains (Section~\ref{subsec:diagnostic}).

\section{Representation-Level Evaluation}
\label{sec:framework}

\subsection{Problem statement}
We formalise RULER in the primary setting of binary classification on tabular data. The training dataset $\mathcal{D}$ consists of records $(x,y)$ with $x\in\reals^d$ and $y\in\{0,1\}$. The predicted class probabilities are $\sigma(f_\theta(x))$, where $\sigma$ is the softmax function and $f_\theta\colon\reals^d\to\reals^2$ is a model with parameters $\theta$ trained on $\mathcal{D}$. The metric definitions are label-free and do not depend on the number of classes; the diagnostic experiments (Section~\ref{subsec:diagnostic}) therefore apply RULER without modification to multi-class image, masked-language-modelling, and face-identity settings, with $f_\theta$ and the embedding dimension differing per architecture (Appendix Table~\ref{tab:setting_summary}).

We assume that the full training set is partitioned into a retain set $\Dr\subseteq\mathcal{D}$ and a forget set $\Df\coloneqq\mathcal{D}\setminus\Dr$, and we define the \textit{forget fraction} $\ff\coloneqq|\Df|/|\mathcal{D}|$. Three models are defined:
\begin{itemize}
  \item the \textit{original} model $\modelO$ trained on $\mathcal{D}$;
  \item an \textit{unlearned} model $\modelU$ produced by applying an unlearning algorithm to $\modelO$;
  \item the \textit{retrain oracle} $\modelR$ trained from scratch on $\Dr$ alone.
\end{itemize}
The oracle serves as the gold-standard reference, a model with no knowledge of $\Df$. Starting from $\modelO$, the goal of unlearning is to produce a model $\modelU$ that behaves as if $\Df$ were never seen, matching the oracle $\modelR$ in outputs and internal representations without the cost of retraining from scratch. Reliable metrics are therefore needed that verify erasure at the representation level, not only at the output level.

\subsection{Penultimate-layer cosine similarity}
The proposed metrics use cosine similarity between penultimate-layer embeddings. Let $h_\theta\colon\reals^d\to\reals^p$ denote the penultimate-layer embedding function of a model $\theta$, i.e.\ the activation immediately before the task-specific output head $g_\theta$, so that $f_\theta = g_\theta\circ h_\theta$. The embedding dimension $p$ is fixed by the architecture; in the primary tabular experiments (Section~\ref{sec:arch}) $p=128$, and the diagnostic experiments use the analogous penultimate activation of each architecture (Section~\ref{subsec:diagnostic}). All embeddings are L2-normalised prior to any similarity computation. 

Thus, for any two records $x,x'\in\reals^d$ and models $\theta,\theta'$, the cosine similarity in the embedding space is:
\begin{equation}
  \similarity_{\theta,\theta'}(x,x') \coloneqq \frac{h_\theta(x)^\top h_{\theta'}(x')}{\lVert h_\theta(x)\rVert\,\lVert h_{\theta'}(x')\rVert} = h_\theta(x)^\top h_{\theta'}(x'),
  \label{eq:cosine}
\end{equation}
where the second equality follows from L2-normalisation. The metrics compare either the same record across two models (\textit{cross-model similarity}, $\similarity_{\theta,\theta'}(x)\coloneqq\similarity_{\theta,\theta'}(x,x)$) or two records within the same model (\textit{within-model similarity}, $\similarity_\theta(x,x')\coloneqq\similarity_{\theta,\theta}(x,x')$).

The penultimate layer is the final feature abstraction before the task-specific output head; earlier layers share lower-level features across records, and the output layer collapses representations into logits or per-token distributions, discarding geometric structure. Cosine similarity operates per record (necessary for small forget sets), is magnitude-invariant in L2-normalised space (enabling cross-dataset comparison), and produces a scalar for statistical testing against a calibrated null. Centred kernel alignment (CKA)~\cite{kornblith2019similarity} and mutual information estimators produce matrix-level summaries that do not naturally decompose per record, making them unsuitable for verifying individual erasure requests under GDPR Article~17.

\subsection{Lens~1: oracle-comparative metrics ($\Mone$, $\Mtwo$, $\Mthree$)}
\label{subsec:lens1}
Lens~1 metrics require access to the retrain oracle $\modelR$ and measure how closely the unlearned model's representations of records in $\Df$ agree with those of the oracle. All Lens~1 metrics use a paired-seed design: the original model $\modelO$ and retrain oracle $\modelR$ share the same random initialisation seed, so that representational differences reflect the unlearning procedure rather than geometric variation across random initialisations.

\paragraph{$\Mone$: oracle-comparison similarity.} For each forget record, $\Mone$ computes the cosine similarity between its embedding under the unlearned model and its embedding under the oracle, then averages these similarities across the forget set:
\begin{equation}
  \Mone = \frac{1}{|\Df|}\sum_{x\in\Df}\similarity_{\modelU,\modelR}(x).
  \label{eq:m1}
\end{equation}
A value near 1.0 indicates that the unlearned model places forget-set records in the same representational neighbourhood as the oracle. $\Mone$ provides an absolute similarity measure but has no fixed reference value against which to test significance, because its expected value under successful unlearning depends on the dataset and random seed. Here and below, we drop the class labels $y$ to simplify the notation, writing `$x \in \mathcal{D}'$' as shorthand for `$(x, y) \in \mathcal{D}'$' for any dataset $\mathcal{D}'$.

\paragraph{$\Mtwo$: signed calibration gap.} To address $\Mone$'s lack of a fixed reference value, $\Mtwo$ subtracts the median cross-model similarity that retain-set records achieve between the same unlearned and oracle model pair, producing a calibrated gap with a null of zero:
\begin{equation}
  \Mtwo = \Mone - \underset{x\in\Dr}{\mathrm{median}}\;\similarity_{\modelU,\modelR}(x).
  \label{eq:m2}
\end{equation}

The median is preferred over the arithmetic mean because the retain-set mean falls below the median by an amount comparable to the gap itself, consistent with a lower tail of retain records that are less well aligned with the oracle.
Replacing the median with the mean baseline shifts $\Mtwo$ by $+0.0023$ for NegGrad$+$ and by $+0.0013$ for Fine-Tuning at $\ff=\SI{1}{\percent}$ (Appendix Table~\ref{tab:m2_baseline_sensitivity}), the latter exceeding the median-baseline gap and reversing its sign. Consequently, replacing the median with the mean baseline reduces significant conditions from $12$ of $12$ to $3$ of $12$ on the pooled Wilcoxon test ($N=100$ per condition) and reverses the sign in $5$ of $12$. The forget set uses the mean ($\Mone$) because it is smaller, and every forget record, including outliers that may signal incomplete erasure, should contribute to the summary. The retain-set median is estimated per seed from a random subsample of $\min\{500,|\Dr|\}$ records. The null hypothesis is $\Mtwo=0$, meaning forget-set records attain the same similarity to the oracle as retained records (oracle-level forgetting); we confirm this null is empirically achievable using independently retrained oracle pairs in Appendix~\ref{app:m2_baseline}. $\Mtwo$ should be interpreted alongside retain accuracy, since a gap of zero is only meaningful if the model remains predictive on $\Dr$. Negative values indicate residual memorisation; positive values indicate the unlearning procedure has moved forget-set representations closer to the oracle than retained records. This is the \textit{primary Lens~1 metric}.

\paragraph{$\Mthree$: representation shift.} $\Mthree$ asks whether unlearning moved forget-set representations closer to or further from the oracle, by comparing each record's unlearned-to-oracle similarity against its original-to-oracle similarity before unlearning:
\begin{equation}
  \Mthree = \frac{1}{|\Df|}\sum_{x\in\Df}\bigl[\similarity_{\modelU,\modelR}(x)-\similarity_{\modelO,\modelR}(x)\bigr].
  \label{eq:m3}
\end{equation}

Positive values indicate that unlearning moved forget-set representations closer to the oracle (the intended outcome), negative values that it moved them further away, and values near zero a negligible effect.

\subsection{Lens~2: oracle-free metric $\Mfour$}
\label{subsec:lens2}
Lens~2 requires no retrained oracle and operates solely on the internal geometry of the unlearned model $\modelU$ and the data partition $\{\Dr,\Df\}$. For each forget record, we find its nearest neighbour in the retain set under the unlearned model and measure their cosine similarity. We then rank this similarity within the leave-one-out nearest-neighbour similarities of the retain records themselves. A forget record that blends in with retained records will sit at the median of this distribution, giving a percentile rank of $0.50$.

Formally, for each forget record $x\in\Df$, define its nearest-retain-neighbour similarity under the unlearned model:
\begin{equation}
  \simf(x) = \max_{x''\in\Dr}\;\similarity_{\modelU}(x,x'').
  \label{eq:sf}
\end{equation}
For each retain record $x'\in\Dr$, define its leave-one-out nearest-retain-neighbour similarity:
\begin{equation}
  \simr(x') = \max_{x''\in\Dr\setminus\{x'\}}\;\similarity_{\modelU}(x',x'').
  \label{eq:sr}
\end{equation}

A retain record would otherwise match itself at similarity $1.0$, which is why $\simr$ takes the maximum over $\Dr \setminus \{x'\}$ rather than all of $\Dr$; a forget record has no such self-match to exclude. Both quantities therefore measure similarity to the closest distinct retain record, so the percentile rank is not inflated by giving forget records an extra candidate neighbour.

\begin{equation}
  \Mfour(x) = \frac{1}{|\Dr|}\sum_{x'\in\Dr}\mathbf{1}\bigl[\simr(x')\leq\simf(x)\bigr],
  \label{eq:m4}
\end{equation}
where $\mathbf{1}[\,\cdot\,]$ is the indicator function. The aggregate $\Mfour$ reported in the results is the mean of $\Mfour(x)$ over all $x\in\Df$.

The null hypothesis is $\Mfour=0.50$. A forget record that has been properly erased should be geometrically indistinguishable from a retained record, sitting at the median of the retain nearest-neighbour distribution. A value above $0.50$ indicates the forget record's nearest-retain-neighbour similarity exceeds that of the median retained record, a sign of residual memorisation. A value below $0.50$ indicates over-displacement, meaning the record has been pushed further from the retain distribution than a correctly retrained model would produce. This is the \textit{primary Lens~2 metric}. 

The use of $\Mfour$ assumes that memorisation is reflected as a geometric deviation in the representation space under the chosen similarity measure. If such deviations are not present, the metric may not detect memorisation.

\paragraph{Relationship between the two lenses.}
$\Mtwo$ measures repositioning relative to the oracle and supports population-level inference across datasets, with negative values indicating residual memorisation, positive values over-correction, and zero values oracle-level erasure. $\Mfour$ measures geometric indistinguishability within the unlearned model alone, making it oracle-free but inherently dataset-specific. We therefore report $\Mfour$ as a per-dataset diagnostic rather than a population-level test.

\section{Experimental Setup}
\label{sec:methodology}

\subsection{Datasets and model architecture}
\label{sec:arch}
Ten tabular classification datasets are used: Adult Income, Diabetes 130-US, Breast Cancer, Heart Disease, German Credit, Bank Marketing, Wine Quality, Phoneme, Magic Telescope, and Electricity. Nine are sourced from OpenML and Breast Cancer from \texttt{sklearn.datasets}; three with multi-class targets are binarised. Each dataset is split \SI{80}{\percent}--\SI{20}{\percent} into training and held-out test partitions (stratified sampling, random state 999); features are standardised using \texttt{StandardScaler} fitted on the training partition.

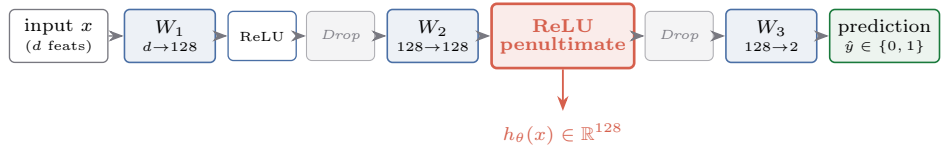
\begin{figure}[ht]
  \centering
  \begin{tikzpicture}[
      font=\scriptsize, every node/.style={align=center},
      input/.style={draw=rulermuted,line width=0.5pt,rounded corners=2pt,fill=white,minimum width=1.3cm,minimum height=0.7cm},
      lin/.style={draw=ruleraccent,line width=0.6pt,rounded corners=2pt,fill=ruleraccent!10,minimum width=1.2cm,minimum height=0.7cm},
      relu/.style={draw=ruleraccent,line width=0.5pt,rounded corners=2pt,fill=white,minimum width=0.9cm,minimum height=0.65cm,font=\tiny},
      drop/.style={draw=rulermuted!60,line width=0.4pt,rounded corners=2pt,fill=rulermuted!8,minimum width=0.9cm,minimum height=0.65cm,font=\tiny\itshape,text=rulermuted},
      pen/.style={draw=forgetred,line width=1.0pt,rounded corners=2pt,fill=forgetred!12,minimum width=1.3cm,minimum height=0.85cm},
      outbox/.style={draw=oraclegreen,line width=0.6pt,rounded corners=2pt,fill=oraclegreen!8,minimum width=1.2cm,minimum height=0.7cm},
      arr/.style={-{Stealth[length=2mm]},rulermuted,line width=0.5pt}]
    \node[input] (x) at (0,0) {input $x$\\[-0.15em]\tiny ($d$ feats)};
    \node[lin, right=0.20cm of x] (W1) {$W_1$\\[-0.15em]\tiny $d{\to}128$};
    \node[relu, right=0.15cm of W1] (r1) {ReLU};
    \node[drop, right=0.12cm of r1] (d1) {Drop};
    \node[lin, right=0.15cm of d1] (W2) {$W_2$\\[-0.15em]\tiny $128{\to}128$};
    \node[pen, right=0.15cm of W2] (pen) {\textbf{\textcolor{forgetred}{ReLU}}\\[-0.15em]\scriptsize\bfseries\color{forgetred}penultimate};
    \node[drop, right=0.12cm of pen] (d2) {Drop};
    \node[lin, right=0.15cm of d2] (W3) {$W_3$\\[-0.15em]\tiny $128{\to}2$};
    \node[outbox, right=0.15cm of W3] (y) {prediction\\[-0.15em]\tiny $\hat y\in\{0,1\}$};
    \foreach \a/\b in {x/W1, W1/r1, r1/d1, d1/W2, W2/pen, pen/d2, d2/W3, W3/y}
      \draw[arr] (\a) -- (\b);
    \draw[forgetred,line width=0.7pt,-{Stealth[length=2mm]}] (pen.south) -- ++(0,-0.55);
    \node[font=\scriptsize\bfseries,text=forgetred,align=center] at ($(pen.south)+(0,-0.85)$) {$h_\theta(x)\in\mathbb{R}^{128}$};
  \end{tikzpicture}
  \caption{The tabular MLP (\texttt{TabularMLP}) used in all primary experiments. RULER extracts the penultimate-layer activation $h_\theta(x)$ (the output of the second ReLU, highlighted in red), L2-normalises it, and feeds it to $\Mtwo$ and $\Mfour$ (Eqs.~\ref{eq:m2}, \ref{eq:m4}). The architecture is held constant across all ten datasets.}
  \label{fig:arch_mlp}
\end{figure}

All models use a two-hidden-layer MLP (\texttt{TabularMLP}) with the fixed architecture shown in Fig.~\ref{fig:arch_mlp}, hidden dimension $p=128$ and dropout 0.2 after each ReLU. The architecture is held constant across all datasets and all three model types ($\modelO$, $\modelU$, $\modelR$). Training uses Adam with learning rate $\eta=10^{-3}$ and cross-entropy loss for 50 epochs full-batch (no mini-batching), chosen to reduce gradient variance and isolate unlearning dynamics; robustness to mini-batch training is assessed in Appendix~\ref{app:minibatch}.

Ten training seeds ($i\in\{0,\ldots,9\}$) are used per dataset; a single unlearning seed ($j=100$) per training seed yields 10 runs per method per dataset per forget fraction and $N=100$ observations (10 datasets~$\times$~10 seeds) per statistical test. For each training seed $i$, $\modelO$ and $\modelR$ are trained with the same random seed (paired-seed design). Three forget fractions are examined: $\ff\in\{\SI{1}{\percent},\SI{5}{\percent},\SI{10}{\percent}\}$. The forget set $\Df$ is constructed by sampling $\max(10,\lfloor\ff\times|\mathcal{D}|\rfloor)$ records without replacement using a fixed random state (999), shared across all experiments. The floor of 10 prevents degenerate forget sets on smaller datasets. The forget-set composition is identical for all four unlearning methods and all ten training seeds within a given dataset and forget fraction. The \SI{1}{\percent} condition is of particular practical relevance, corresponding to the typical scale of individual erasure requests under GDPR right-to-erasure provisions. Robustness to forget-set sampling is verified in Appendix~\ref{app:forget_seed}.

\subsection{Unlearning methods}
\label{sec:unlearn_methods}
\textbf{Gradient Ascent (GA)} maximises cross-entropy loss on $\Df$ for 5 epochs. \textbf{NegGrad$+$ (NG$+$)} combines gradient ascent on $\Df$ with gradient descent on $\Dr$ for 10 epochs, weighted by $\alpha=0.6$. \textbf{Fine-Tuning (FT)} continues gradient descent exclusively on $\Dr$ for 10 epochs, relying on catastrophic forgetting. \textbf{SCRUB} distils from the frozen original (teacher) to a student copy for 10 epochs, maximising agreement on $\Dr$ and disagreement on $\Df$~\cite{kurmanji2023towards}, with $\alpha=0.6$ and temperature $T=2.0$. All four start from $\modelO$ and use Adam with unlearning learning rate $\eta_u=5\times 10^{-4}$, held fixed across all datasets and forget fractions. A fifth method, Bad Teacher~\cite{chundawat2023can}, is evaluated separately (Section~\ref{subsec:bad_teacher_body}; implementation in Appendix~\ref{app:bad_teacher}).

\subsection{Evaluation protocol and statistical analysis}
\label{sec:stats}
The oracle-comparative metrics operate on penultimate-layer embeddings; for the retain-set baseline in $\Mtwo$, a subsample of $\min(500,|\Dr|)$ records (random state 42) is drawn per seed. $\Mfour$ computes nearest-retain-neighbour similarities across all $\Dr$ records, capped at 2{,}000 for memory. Output-level metrics are computed in parallel: MIA accuracy (threshold-based attack using per-sample cross-entropy loss, with balanced accuracy reported), retain, forget, and test accuracy. Let $\overline{\text{MIA}}$ denote the mean MIA accuracy across the 10 training seeds per (dataset, method, forget fraction) condition. The pass threshold $|\overline{\text{MIA}}-0.50|<0.05$ accommodates dataset-level variation.

We test $\Mtwo$ against null $0$ and $\Mfour$ against null $0.50$. The primary analysis is a linear mixed-effects model (LMM) with random intercept for dataset, fitted via restricted maximum likelihood (REML) to all $N=100$ observations; the fixed intercept is tested using the Wald $z$-statistic. With only 10 clusters this test can overstate significance, so we also report the Wilcoxon signed-rank test on $N=10$ dataset-level means. The intra-class correlation $\mathrm{ICC}=\sigma_u^2/(\sigma_u^2+\sigma_\varepsilon^2)$, where $\sigma_u^2$ is the between-dataset (random-intercept) variance and $\sigma_\varepsilon^2$ the within-dataset (residual) variance estimated by the LMM, quantifies between-dataset variance share. Effect sizes are reported as rank-biserial correlation $\rrb = 1 - 4W/[n(n+1)]$, where $W$ is the Wilcoxon statistic and $n$ the number of non-tied observations; $|\rrb|\geq 0.10, 0.30, 0.50$ denote small, medium, large effects per conventional thresholds.

\section{Results}
\label{sec:results}
\noindent\textbf{Key findings.} All four methods pass output-level evaluation (Appendix Table~\ref{tab:output_level}), but $\Mtwo$ detects significant residuals in 10 of 12 conditions under the LMM (Table~\ref{tab:cross_ff}), with all four methods significant at the highest forget fraction and directional consistency across datasets (Appendix Fig.~\ref{fig:per_dataset_shift}). After correction, methods are statistically indistinguishable on $\Mtwo$, $\Mfour$, and MIA, indicating the discordance is a property of the task, not any algorithm. Bad Teacher reproduces the same negative $\Mtwo$ at every fraction despite a fundamentally different mechanism (Section~\ref{subsec:bad_teacher_body}). $\Mfour$ is dataset-specific (ICC $=0.89$) and serves as a per-dataset diagnostic, sensitive to identity-level memorisation (Appendix Fig.~\ref{fig:lfw}) and over-displacement on deeper architectures.

\subsection{Representation-level residuals persist despite output-level pass}
\label{subsec:residuals}
All four methods pass standard output-level evaluation at every forget fraction, with mean MIA accuracy within $\pm 0.05$ of $0.50$ and retain-set and test-set accuracy preserved (Appendix Table~\ref{tab:output_level}). Per-dataset MIA results show variation masked by the aggregate (Appendix Table~\ref{tab:perdataset_mia}). On German Credit at $\ff=\SI{1}{\percent}$, the oracle itself breaches the $\pm 0.05$ window (MIA $=0.700$), indicating a dataset-level characteristic rather than incomplete forgetting; Breast Cancer ($0.560$) similarly breaches at the oracle level, whilst Heart Disease shows method-specific breaches for NegGrad$+$ ($0.450$), Fine-Tuning, and SCRUB ($0.570$; oracle $=0.520$). Despite passing all output-level checks on aggregate, the oracle-comparative metric $\Mtwo$ detects residuals in 10 of 12 method-fraction conditions ($p<0.05$, LMM; Table~\ref{tab:cross_ff}), confirming that output-level evaluation is insufficient.

\begin{table}[t]\footnotesize
\caption{$\Mtwo$ and $\Mfour$ across all three forget fractions ($N=100$ per method; 10 datasets~$\times$~10 seeds). $p$-values from a linear mixed-effects model with random intercept for dataset; effect sizes $\rrb$ are rank-biserial correlations from the pooled Wilcoxon signed-rank test ($N=100$). $\Mfour$ does not reach significance under the LMM at any forget fraction (ICC $=0.89$); see Section~\ref{subsec:diagnostic} for its role as a per-dataset diagnostic. Significance: $^*p<0.05$, $^{**}p<0.01$, $^{***}p<0.001$.}
\label{tab:cross_ff}
\centering
\setlength{\tabcolsep}{6pt}
\begin{tabular}{lcccccc}
\toprule
 & \multicolumn{3}{c}{$\Mtwo$} & \multicolumn{3}{c}{$\Mfour$} \\
\cmidrule(lr){2-4}\cmidrule(lr){5-7}
Method & gap & $p$ & $\rrb$ & rank & $p$ & $\rrb$ \\
\midrule
\multicolumn{7}{l}{\textbf{$\ff = \SI{1}{\percent}$}} \\
Gradient Ascent & $-0.00077$ & $0.136$         & $+0.51$ & $0.533$ & $0.082$ & $+0.47$ \\
NegGrad$+$      & $-0.00470$ & $0.008^{\mathrlap{**}}$    & $+0.87$ & $0.532$ & $0.100$ & $+0.43$ \\
Fine-Tuning     & $-0.00112$ & $0.043^{\mathrlap{*}}$     & $+0.71$ & $0.538$ & $0.053$ & $+0.52$ \\
SCRUB           & $-0.00121$ & $0.024^{\mathrlap{*}}$     & $+0.71$ & $0.538$ & $0.052$ & $+0.52$ \\
\addlinespace
\multicolumn{7}{l}{\textbf{$\ff = \SI{5}{\percent}$}} \\
Gradient Ascent & $-0.00097$ & $0.064$         & $+0.54$ & $0.518$ & $0.114$ & $+0.43$ \\
NegGrad$+$      & $-0.00203$ & $0.003^{\mathrlap{**}}$    & $+0.76$ & $0.518$ & $0.114$ & $+0.44$ \\
Fine-Tuning     & $-0.00134$ & $0.035^{\mathrlap{*}}$     & $+0.73$ & $0.522$ & $0.078$ & $+0.52$ \\
SCRUB           & $-0.00122$ & $0.045^{\mathrlap{*}}$     & $+0.71$ & $0.521$ & $0.083$ & $+0.51$ \\
\addlinespace
\multicolumn{7}{l}{\textbf{$\ff = \SI{10}{\percent}$}} \\
Gradient Ascent & $-0.00167$ & $\mathllap{<}0.001^{\mathrlap{***}}$  & $+0.90$ & $0.506$ & $0.357$ & $+0.25$ \\
NegGrad$+$      & $-0.00171$ & $\mathllap{<}0.001^{\mathrlap{***}}$  & $+0.96$ & $0.506$ & $0.330$ & $+0.26$ \\
Fine-Tuning     & $-0.00210$ & $0.004^{\mathrlap{**}}$    & $+0.93$ & $0.506$ & $0.321$ & $+0.22$ \\
SCRUB           & $-0.00207$ & $0.003^{\mathrlap{**}}$    & $+0.94$ & $0.506$ & $0.346$ & $+0.21$ \\
\bottomrule
\end{tabular}
\end{table}

The oracle-comparative metric $\Mtwo$ is negative for every method at every forget fraction (Table~\ref{tab:cross_ff}), meaning that after unlearning, forget-set records have lower cosine similarity to the oracle than retained records. Under the LMM, 10 of 12 method-fraction conditions are significant at $p<0.05$, with the two non-significant conditions (Gradient Ascent at $\ff=\SI{1}{\percent}$, $p=0.136$, and $\ff=\SI{5}{\percent}$, $p=0.064$) remaining directionally consistent. 

The oracle-free metric $\Mfour$ shows mean ranks above $0.50$ in all 12 conditions. Its variance is dominated by dataset identity (ICC $=0.89$), which is why it functions as a per-dataset diagnostic rather than a population-level test. The per-dataset analysis shows that a small number of datasets with elevated $\Mfour$ (e.g.\ Breast Cancer, $\Mfour\approx 0.60$; Heart Disease, $\Mfour\approx 0.55$) contribute most of the aggregate deviation from $0.50$, whilst the remaining datasets sit near the null for both unlearned and oracle models. These gaps are small in absolute terms because L2-normalised penultimate embeddings place all records near similarity $1.0$ under the paired-seed design, so the calibrated gap rather than the raw similarity carries the signal. They are nonetheless consistent in sign across methods, datasets, and forget fractions (Appendix Fig.~\ref{fig:per_dataset_shift}). Eight of ten datasets show a negative representation shift for all four methods at $\ff=\SI{5}{\percent}$.

Heart Disease and German Credit are exceptions at $\ff=\SI{5}{\percent}$, consistent with their small forget sets ($|\Df|=12$ and $40$) limiting per-dataset power.

%The complementary metric $\Mthree$ (Eq.~\ref{eq:m3}) shows that all four methods produce negative representation shifts after unlearning, meaning unlearning moves forget-set representations further from the oracle rather than closer.

Aggregated across datasets, the complementary metric $\Mthree$ (Eq.~\ref{eq:m3}) shows that all four methods produce negative representation shifts after unlearning, meaning unlearning moves forget-set representations further from the oracle rather than closer.

\subsection{Metric behaviour across forget fractions}
\label{subsec:metric_divergence}
$\Mtwo$ effect sizes increase with the forget fraction, reaching their maximum at $\ff=\SI{10}{\percent}$ ($\rrb=0.90$ to $0.96$), with all four methods significant at $p<0.005$ under the LMM (Table~\ref{tab:cross_ff}). $\Mfour$ does not reach population-level significance at any fraction (ICC $=0.89$), with effect sizes decreasing from medium-to-large at $\ff=\SI{1}{\percent}$ to small at $\ff=\SI{10}{\percent}$. The per-dataset distribution of $\Mfour$ (Appendix Fig.~\ref{fig:per_dataset_m4}) shows that within each dataset the four methods and the oracle baseline produce similar distributions; deviations from $0.50$ therefore reflect feature-space geometry rather than residual memorisation specific to the unlearning algorithm, consistent with the framing of $\Mfour$ as a per-dataset diagnostic. The residual signal also holds across unlearning learning rates: $\Mtwo<0$ and $\Mfour>0.50$ for the same datasets at $\eta_u\in\{10^{-4},5\times 10^{-4},10^{-3}\}$ (Appendix~\ref{app:sensitivity}, Table~\ref{tab:sensitivity}).

\subsection{Methods are interchangeable on representation metrics but differ on retain accuracy}
\label{subsec:method_comparison}
Post-hoc pairwise Wilcoxon tests with Benjamini--Hochberg correction (six method pairs, four metrics, 24 tests) distinguish no pair on $\Mtwo$, $\Mfour$, or MIA accuracy ($p_{\mathrm{adj}}>0.05$): the discordance is a property of the unlearning task, not of any individual algorithm. Fine-Tuning and SCRUB preserve retain-set performance significantly better than Gradient Ascent and NegGrad$+$ ($p_{\mathrm{adj}}=0.046$ for all four cross-group pairs), but this advantage does not reduce $\Mtwo$ or $\Mfour$ residuals.

\subsection{Findings hold under a different mechanism: Bad Teacher}
\label{subsec:bad_teacher_body}
\begin{figure}[ht]
  \centering
  \includegraphics[width=\linewidth]{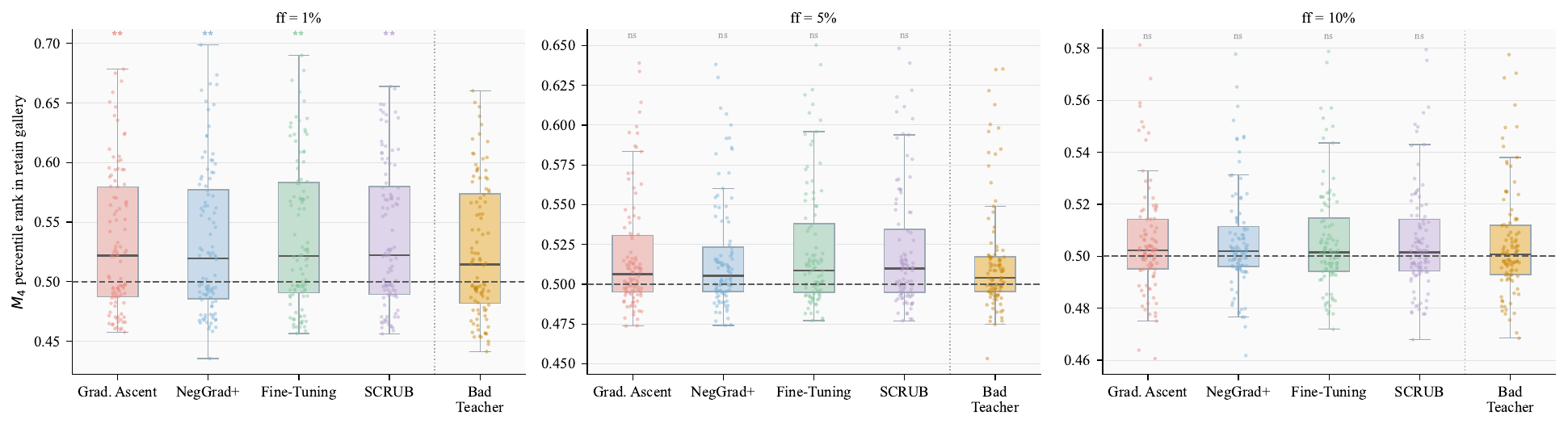}
\caption{Oracle-free percentile rank $\Mfour$ for the four approximate methods and Bad Teacher across all forget fractions ($N=100$ per method). Dashed line: null of $0.50$. Bad Teacher displaces forget-set records furthest from the retain set at $\ff=\SI{1}{\percent}$, whilst its $\Mtwo$ remains significantly negative throughout (Appendix Table~\ref{tab:bt_repr}). Markers: paired Wilcoxon tests against Bad Teacher (Appendix Table~\ref{tab:bt_paired}); $^{**}p\leq 0.01$, ns $p\geq 0.05$.}
  \label{fig:bt_m4}
\end{figure}

To check whether the discordance is method-specific or structural, we evaluate a fifth method, Bad Teacher~\cite{chundawat2023can}, which trains a student to match a randomly initialised teacher on forget-set records (Appendix~\ref{app:bad_teacher}). Bad Teacher passes output-level evaluation at all three forget fractions yet shows significant $\Mtwo$ residuals throughout ($p<0.01$; Appendix Table~\ref{tab:bt_repr}), and on $\Mfour$ (Fig.~\ref{fig:bt_m4}) displaces forget-set records further from the retain set than all four approximate methods at $\ff=\SI{1}{\percent}$ without achieving complete erasure. The discordance therefore holds across five methods spanning direct parameter modification and distillation.

\subsection{Diagnostic role of $\Mfour$ beyond tabular MLPs}
\label{subsec:diagnostic}
Beyond verifying unlearning, $\Mfour$ can indicate whether unlearning is necessary at the representation level. Applied to the original model, it reveals whether forget-set records are memorised or blend into the general distribution; when memorisation is absent, applying unlearning may introduce artefacts. We evaluate this diagnostic across four settings: large-scale tabular data (Covertype, Higgs; 100{,}000 samples; MLP, Residual MLP, FT-Transformer), image classification (CIFAR-10, SVHN, CIFAR-100; three-layer CNN and ResNet-18), clinical text (MTSamples with Bio\_ClinicalBERT, masked language modelling), and face-identity unlearning (LFW; ResNet-18 trained for gender classification). Architecture diagrams are in Appendix Figs.~\ref{fig:arch_image} and~\ref{fig:arch_bert}.

\paragraph{Large-scale tabular datasets.}
Across Covertype (forest cover type, 7 classes) and Higgs (binary signal-versus-background; 100{,}000 samples each), using MLP, Residual MLP, and FT-Transformer with mini-batch training (batch size 256, 5 seeds, $\ff=\SI{5}{\percent}$), original-model $\Mfour$ values are close to $0.50$ for all three architectures (Appendix Table~\ref{tab:largescale}), indicating weak or no detectable memorisation. Unlearning has little effect, with most methods remaining near $0.50$. NegGrad$+$ is the exception, increasing $\Mfour$ (e.g.\ $0.60$ on Higgs with FT-Transformer), indicating displacement rather than removal. In contrast to the primary tabular experiment, no memorisation is detected at this scale, and when memorisation is weak or absent, unlearning methods can introduce artefacts.

\paragraph{Image data.}
We repeated the analysis on CIFAR-10, SVHN, and CIFAR-100 (10{,}000 samples each, $\ff=\SI{5}{\percent}$) with a three-layer CNN and ResNet-18. Original models again show $\Mfour\approx 0.50$ (Appendix Table~\ref{tab:image_m4}), indicating no or weak detectable memorisation. After unlearning, behaviour depends on model capacity. On ResNet-18, NegGrad$+$ and SCRUB produce substantial over-displacement ($\Mfour=0.03$ to $0.16$ and $0.11$ to $0.37$ respectively); on the three-layer CNN most methods remain near $0.50$. Some methods therefore introduce representation-level artefacts when memorisation is weak or absent, particularly in higher-capacity models.

\begin{figure}[t]
  \centering
  \includegraphics[width=0.95\linewidth]{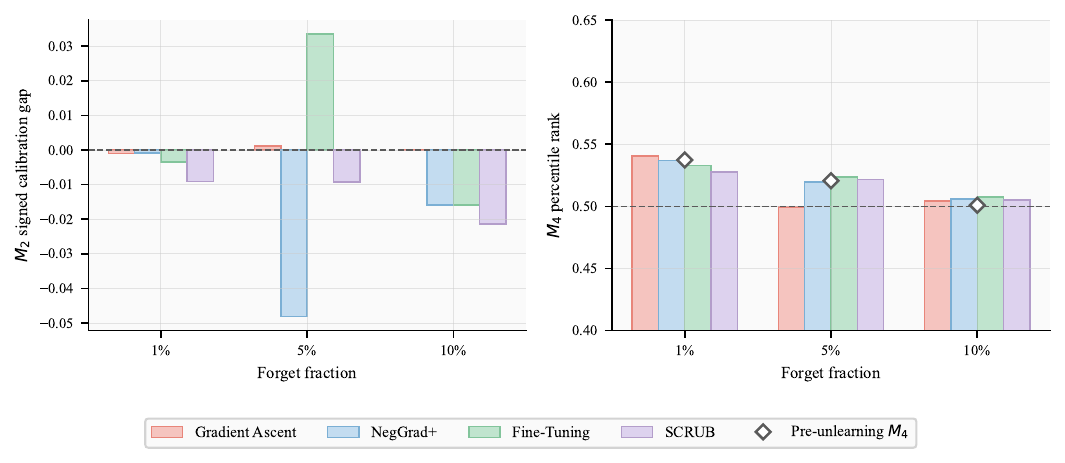}
  \caption{RULER on Bio\_ClinicalBERT (clinical text). (a)~$\Mtwo$ remains small at \SI{1}{\percent} but grows more variable at larger forget fractions. (b)~$\Mfour$ stays near its $0.50$ null across fractions, indicating weak representation-level memorisation. Dashed lines mark the nulls; the diamond marks the pre-unlearning $\Mfour$ diagnostic.}
  \label{fig:clinical_m4}

\end{figure}

\paragraph{Clinical text.}
We next tested whether RULER applies to a non-classification training objective. Bio\_ClinicalBERT is a BERT-base model pre-trained on clinical notes using masked language modelling. We evaluated it on MTSamples using CLS-token representations, with forget and retain sets defined at the document level and expanded to sentences (Fig.~\ref{fig:clinical_m4}). The pre-unlearning $\Mfour$ diagnostic showed mean values of $0.537$, $0.521$, $0.501$ at $\ff=\SI{1}{\percent}$, \SI{5}{\percent}, \SI{10}{\percent}. The \SI{5}{\percent} and \SI{10}{\percent} values indicate no detectable memorisation, and the \SI{1}{\percent} value sits only slightly above the null. After unlearning, $\Mtwo$ remains small at \SI{1}{\percent} for most methods but shows increased variability at larger forget fractions, with both negative and positive values observed (e.g.\ $-0.048$ for NegGrad$+$ and $+0.033$ for Fine-Tuning at \SI{5}{\percent}).

\paragraph{Face identity unlearning.}
Erasure requests under GDPR target specific individuals, with the forget set comprising all records belonging to one person. To test whether $\Mfour$ detects identity-level memorisation, we trained a ResNet-18 on gender classification using Labeled Faces in the Wild (LFW; 4{,}324 images, 158 identities, 5 seeds), so that identity is incidental to the training objective; forget sets comprise all images of target identities at each $\ff$.
Pre-unlearning $\Mfour$ is $0.94$, $0.77$, $0.73$ at $\ff=\SI{1}{\percent}$, \SI{5}{\percent}, \SI{10}{\percent} (Appendix Fig.~\ref{fig:lfw} and Table~\ref{tab:lfw_m4}), all far above the null, in contrast to CIFAR/SVHN. After unlearning, Gradient Ascent and Fine-Tuning leave identity structure largely intact ($\Mfour=0.70$ to $0.91$), NegGrad$+$ partially reduces it, and SCRUB over-displaces at \SI{1}{\percent} ($\Mfour=0.24$). $\Mtwo$ is close to zero with high variance, so $\Mfour$ carries the residual signal. Across the four diagnostic settings, weak memorisation leaves $\Mfour$ near its null, where unlearning has limited effect or may introduce artefacts; when strong, no tested method returns $\Mfour$ to the null.

\section{Discussion and Conclusion}
\label{sec:discussion}
All four approximate methods pass output-level evaluation but leave statistically significant $\Mtwo$ residuals, and Bad Teacher reproduces the same outcome despite a different forgetting mechanism, so the discordance is a property of the unlearning task, not of any individual algorithm. Fine-Tuning and SCRUB better preserve retain-set performance, but this advantage does not extend to the representation-level metrics, and NegGrad$+$ and SCRUB introduce representation-level artefacts on deeper architectures. Before unlearning, $\Mfour$ values near $0.50$ indicate that unlearning may be unnecessary, whilst values far above it ($0.73$ to $0.94$ on LFW) confirm identity-level structure, and no tested method returns $\Mfour$ to the null. After unlearning, deviations from the null flag incomplete erasure or over-displacement that output-level metrics miss. The tabular residuals are small ($|\Mtwo|\approx 10^{-3}$) and for a black-box classifier we show no exploitable risk; their significance is that output-level evaluation certifies erasure that remains incomplete. Where embeddings are exposed, through APIs, transfer learning, or retrieval, representational proximity of forget-set records supports membership inference, and where the residual is identity-linked, as on LFW, the risk is re-identification. $\Mtwo$ requires a retrain oracle, the paired-seed design, and full-batch training (Appendix~\ref{app:minibatch}); the oracle costs exactly the retraining unlearning avoids, so $\Mtwo$ suits audits, not per-request checks or foundation-scale models. $\Mfour$ needs only the unlearned model and the data partition and persists under mini-batch training, although it is dataset-specific (ICC $=0.89$). Future unlearning objectives should incorporate representation-level constraints directly. Methodological limitations are detailed in Appendix~\ref{app:limitations}.

\subsubsection*{Disclosure of Interests.} The authors have no competing interests to declare.

\bibliographystyle{splncs04}
\bibliography{references}

\begin{thebibliography}{10}
\providecommand{\url}[1]{\texttt{#1}}
\providecommand{\urlprefix}{URL }
\providecommand{\doi}[1]{https://doi.org/#1}

\bibitem{bourtoule2021machine}
Bourtoule, L., Chandrasekaran, V., Choquette-Choo, C.A., Jia, H., Travers, A., Zhang, B., Lie, D., Papernot, N.: Machine unlearning. In: 2021 IEEE Symposium on Security and Privacy (SP). pp. 141--159. IEEE (2021)

\bibitem{cao2015towards}
Cao, Y., Yang, J.: Towards making systems forget with machine unlearning. In: 2015 IEEE Symposium on Security and Privacy. pp. 463--480. IEEE (2015)

\bibitem{chundawat2023can}
Chundawat, V.S., Tarun, A.K., Mandal, M., Kankanhalli, M.S.: Can bad teaching induce forgetting? unlearning in deep networks using an incompetent teacher. In: {AAAI}. pp. 7210--7217. {AAAI} Press (2023)

\bibitem{cf-k}
Goel, S., Prabhu, A., Sanyal, A., Lim, S.N., Torr, P., Kumaraguru, P.: Towards adversarial evaluations for inexact machine unlearning (2023), \url{https://arxiv.org/abs/2201.06640}

\bibitem{golatkar2020eternal}
Golatkar, A., Achille, A., Soatto, S.: Eternal sunshine of the spotless net: Selective forgetting in deep networks. In: Proceedings of the IEEE/CVF Conference on Computer Vision and Pattern Recognition. pp. 9304--9312 (2020)

\bibitem{golatkar2020forgetting}
Golatkar, A., Achille, A., Soatto, S.: {Forgetting Outside the Box}: Scrubbing deep networks of information accessible from input-output observations. In: European Conference on Computer Vision. pp. 383--398 (2020)

\bibitem{hayes2024inexact}
Hayes, J., Shumailov, I., Triantafillou, E., Khalifa, A., Papernot, N.: Inexact unlearning needs more careful evaluations to avoid a false sense of privacy (2024)

\bibitem{kornblith2019similarity}
Kornblith, S., Norouzi, M., Lee, H., Hinton, G.: Similarity of neural network representations revisited. In: Chaudhuri, K., Salakhutdinov, R. (eds.) Proceedings of the 36th International Conference on Machine Learning. Proceedings of Machine Learning Research, vol.~97, pp. 3519--3529. PMLR (09--15 Jun 2019)

\bibitem{kurmanji2023towards}
Kurmanji, M., Triantafillou, P., Hayes, J., Triantafillou, E.: Towards unbounded machine unlearning. In: Thirty-seventh Conference on Neural Information Processing Systems (2023)

\bibitem{shokri2017membership}
Shokri, R., Stronati, M., Song, C., Shmatikov, V.: Membership inference attacks against machine learning models. In: 2017 IEEE Symposium on Security and Privacy (SP). pp. 3--18. IEEE (2017)

\bibitem{thudi2022auditable}
Thudi, A., Jia, H., Shumailov, I., Papernot, N.: On the necessity of auditable algorithmic definitions for machine unlearning. In: 31st {USENIX} Security Symposium ({USENIX} Security 22) (2022)

\bibitem{Wang2025EMUModelDifference}
Wang, W., Zhang, C., Tian, Z., Yu, S., Su, Z.: Evaluation of machine unlearning through model difference. IEEE Transactions on Information Forensics and Security  \textbf{20},  5211--5223 (2025)

\end{thebibliography}

\appendix
\raggedbottom
\section{Additional Multi-Seed Experiments and Robustness Checks}
\label{app:robust}

\subsection{Experimental settings summary}
\label{app:settings}

\begin{table}[H]
\footnotesize
\centering
\caption{Experimental settings and representation extraction across the paper. In all settings, RULER operates on the activation immediately before the task-specific output head: the second ReLU for the Tabular MLP; the final hidden layer for the Residual MLP and FT-Transformer; the second fully-connected layer for the three-layer CNN; the post-global-average-pooling activation for ResNet-18; and the CLS-token output of the final transformer layer for Bio\_ClinicalBERT. The output dimension is the number of classes in every setting except Bio\_ClinicalBERT, which uses masked language modelling, where it is the vocabulary size.}
\label{tab:setting_summary}
\setlength{\tabcolsep}{6pt}
\begin{tabular}{llcr}
\toprule
Dataset(s) & Architecture & Output dim. & Emb.\ dim. \\
\midrule
\multicolumn{4}{l}{\textbf{Primary tabular evaluation (\S\ref{sec:methodology})}} \\
10 OpenML datasets & Tabular MLP & 2 & 128 \\
\addlinespace
\multicolumn{4}{l}{\textbf{Diagnostic: large-scale tabular (\S\ref{subsec:diagnostic})}} \\
Covertype & MLP / Res.\ MLP / FT-Trans. & 7 & 128 \\
Higgs & MLP / Res.\ MLP / FT-Trans. & 2 & 128 \\
\addlinespace
\multicolumn{4}{l}{\textbf{Diagnostic: image (\S\ref{subsec:diagnostic})}} \\
CIFAR-10, SVHN & Three-layer CNN & 10 & 256 \\
CIFAR-10, SVHN & ResNet-18 & 10 & 512 \\
CIFAR-100 & Three-layer CNN & 100 & 256 \\
CIFAR-100 & ResNet-18 & 100 & 512 \\
\addlinespace
\multicolumn{4}{l}{\textbf{Diagnostic: clinical text (\S\ref{subsec:diagnostic})}} \\
MTSamples & Bio\_ClinicalBERT & $\sim$30{,}000 & 768 \\
\addlinespace
\multicolumn{4}{l}{\textbf{Diagnostic: face identity (\S\ref{subsec:diagnostic})}} \\
LFW & ResNet-18 & 2 & 512 \\
\bottomrule
\end{tabular}
\end{table}

\subsection{Methodological limitations}
\label{app:limitations}
The following limitations and assumptions apply to the framework presented in the main text. (1)~The metrics are technical verification tools rather than legal compliance tests under GDPR Article~17; they characterise representational residuals, not regulatory adequacy. (2)~Cosine similarity is not rotation-invariant, so $\Mtwo$ requires the paired-seed design described in Section~\ref{subsec:lens1} to remain meaningful. (3)~Diagnostic experiments (large-scale tabular, image, clinical text, LFW) use five seeds per condition, leaving open whether the over-displacement observed for NegGrad$+$ and SCRUB on deeper architectures can be mitigated by hyperparameter tuning. (4)~A single unlearning seed is used per training seed to separate variability from training initialisation from that of the unlearning procedure; sensitivity to the unlearning seed is untested for the four main methods, though sensitivity to forget-set sampling is characterised across five additional seeds (Appendix~\ref{app:forget_seed}, Table~\ref{tab:forget_seed_sensitivity}) and Bad Teacher is stable across three teacher seeds (Appendix~\ref{app:bad_teacher}). (5)~The primary tabular experiments use uniformly random forget sets, and the LFW experiment uses structured identity-level forget sets; broader coverage of structured deletion scenarios (e.g.\ demographic groups, temporal cohorts) remains an open direction.

\subsection{Output-level evaluation summary}
\label{app:output}
Table~\ref{tab:output_level} reports the full output-level evaluation summary for the four primary unlearning methods. All four methods pass the output-level pass criterion ($|\overline{\text{MIA}}-0.50|<0.05$) at all three forget fractions, with retain and test accuracy preserved at oracle levels.

\begin{table}[H]\footnotesize
\caption{Output-level evaluation summary (mean $\pm$ SD, $N=100$ per method; 10 datasets~$\times$~10 seeds). MIA null $=0.50$. Oracle MIA: $0.530$ at $\ff=\SI{1}{\percent}$; $0.507$ at $\ff=\SI{5}{\percent}$; $0.489$ at $\ff=\SI{10}{\percent}$. All methods pass the output-level criterion ($|\overline{\text{MIA}}-0.50|<0.05$) at all three forget fractions.}
\label{tab:output_level}
\centering
\setlength{\tabcolsep}{4pt}
\begin{tabular}{llcccc}
\toprule
$\ff$ & Method & MIA acc. & Forget acc. & Retain acc. & Test acc. \\
\midrule
\multirow{4}{*}{\SI{1}{\percent}}
    & Gradient Ascent & $0.513 \pm 0.078$ & $0.849 \pm 0.068$ & $0.838 \pm 0.068$ & $0.826 \pm 0.064$ \\
    & NegGrad$+$      & $0.499 \pm 0.063$ & $0.815 \pm 0.063$ & $0.836 \pm 0.068$ & $0.824 \pm 0.065$ \\
    & Fine-Tuning     & $0.536 \pm 0.066$ & $0.877 \pm 0.071$ & $0.848 \pm 0.066$ & $0.836 \pm 0.066$ \\
    & SCRUB           & $0.539 \pm 0.068$ & $0.878 \pm 0.070$ & $0.848 \pm 0.066$ & $0.836 \pm 0.066$ \\
\midrule
\multirow{4}{*}{\SI{5}{\percent}}
    & Gradient Ascent & $0.500 \pm 0.037$ & $0.817 \pm 0.089$ & $0.837 \pm 0.068$ & $0.824 \pm 0.065$ \\
    & NegGrad$+$      & $0.494 \pm 0.041$ & $0.800 \pm 0.097$ & $0.835 \pm 0.070$ & $0.822 \pm 0.068$ \\
    & Fine-Tuning     & $0.513 \pm 0.030$ & $0.849 \pm 0.067$ & $0.848 \pm 0.065$ & $0.836 \pm 0.066$ \\
    & SCRUB           & $0.512 \pm 0.034$ & $0.849 \pm 0.067$ & $0.847 \pm 0.065$ & $0.836 \pm 0.065$ \\
\midrule
\multirow{4}{*}{\SI{10}{\percent}}
    & Gradient Ascent & $0.491 \pm 0.030$ & $0.809 \pm 0.085$ & $0.838 \pm 0.070$ & $0.823 \pm 0.069$ \\
    & NegGrad$+$      & $0.486 \pm 0.038$ & $0.801 \pm 0.097$ & $0.838 \pm 0.070$ & $0.822 \pm 0.072$ \\
    & Fine-Tuning     & $0.493 \pm 0.025$ & $0.836 \pm 0.073$ & $0.850 \pm 0.065$ & $0.835 \pm 0.068$ \\
    & SCRUB           & $0.492 \pm 0.027$ & $0.834 \pm 0.072$ & $0.849 \pm 0.065$ & $0.835 \pm 0.067$ \\
\bottomrule
\end{tabular}
\end{table}

\subsection{Per-dataset MIA breakdown}
\label{app:perdataset_mia}
Table~\ref{tab:perdataset_mia} reports per-dataset MIA accuracy. In most cases where a method breaches the $\pm 0.05$ pass window, the retrained oracle deviates in the same direction, indicating a dataset-level characteristic such as class imbalance or skewed loss distributions rather than incomplete forgetting. The exceptions are, at $\ff=\SI{1}{\percent}$, NegGrad$+$ on Wine Quality ($0.425$; oracle $=0.500$) and Heart Disease, where NegGrad$+$ breaches low ($0.450$) and Fine-Tuning and SCRUB breach high ($0.570$; oracle $=0.520$), and, at $\ff=\SI{5}{\percent}$, Fine-Tuning and SCRUB on German Credit ($0.555$; oracle $=0.537$). In each case the oracle remains within the pass window. These method-specific deviations are isolated cases; the aggregate output-level conclusion (all four methods pass) holds, while $\Mtwo$ detects representation-level residuals consistently across datasets and methods.

\begin{table}[H]
\footnotesize
\centering
\caption{Per-dataset MIA accuracy means across all three forget fractions ($N=10$ seeds per dataset). Bold values breach the $\pm 0.05$ pass window ($|\overline{\text{MIA}}-0.50|\geq 0.05$).}
\label{tab:perdataset_mia}
\setlength{\tabcolsep}{4pt}
\begin{tabular}{rlccccc}
\toprule
$\ff$ & Dataset & Grad.\ Ascent & NegGrad$+$ & Fine-Tuning & SCRUB & Oracle \\
\midrule
\multirow{10}{*}{\SI{1}{\percent}} & Adult Income & $0.489$ & $0.489$ & $0.494$ & $0.494$ & $0.493$ \\
 & Diabetes 130-US & $0.515$ & $0.511$ & $0.516$ & $0.516$ & $0.519$ \\
 & Breast Cancer & $0.480$ & $0.470$ & $\mathbf{0.560}$ & $\mathbf{0.570}$ & $\mathbf{0.560}$ \\
 & Heart Disease & $0.480$ & $\mathbf{0.450}$ & $\mathbf{0.570}$ & $\mathbf{0.570}$ & $0.520$ \\
 & German Credit & $\mathbf{0.720}$ & $\mathbf{0.640}$ & $\mathbf{0.700}$ & $\mathbf{0.710}$ & $\mathbf{0.700}$ \\
 & Bank Marketing & $0.488$ & $0.489$ & $0.494$ & $0.493$ & $0.491$ \\
 & Wine Quality & $0.458$ & $\mathbf{0.425}$ & $0.500$ & $0.508$ & $0.500$ \\
 & Phoneme & $0.474$ & $0.488$ & $0.493$ & $0.493$ & $0.486$ \\
 & Magic Telescope & $0.511$ & $0.518$ & $0.513$ & $0.514$ & $0.513$ \\
 & Electricity & $0.517$ & $0.513$ & $0.521$ & $0.520$ & $0.516$ \\
\midrule
\multirow{10}{*}{\SI{5}{\percent}} & Adult Income & $0.501$ & $0.503$ & $0.505$ & $0.505$ & $0.504$ \\
 & Diabetes 130-US & $0.508$ & $0.507$ & $0.511$ & $0.511$ & $0.510$ \\
 & Breast Cancer & $0.518$ & $0.500$ & $\mathbf{0.564}$ & $\mathbf{0.568}$ & $\mathbf{0.568}$ \\
 & Heart Disease & $\mathbf{0.425}$ & $\mathbf{0.392}$ & $0.483$ & $0.475$ & $\mathbf{0.450}$ \\
 & German Credit & $0.540$ & $0.532$ & $\mathbf{0.555}$ & $\mathbf{0.555}$ & $0.537$ \\
 & Bank Marketing & $0.508$ & $0.508$ & $0.508$ & $0.507$ & $0.509$ \\
 & Wine Quality & $0.483$ & $0.478$ & $0.486$ & $0.481$ & $0.479$ \\
 & Phoneme & $0.512$ & $0.511$ & $0.509$ & $0.508$ & $0.509$ \\
 & Magic Telescope & $0.499$ & $0.498$ & $0.500$ & $0.500$ & $0.497$ \\
 & Electricity & $0.509$ & $0.512$ & $0.511$ & $0.510$ & $0.508$ \\
\midrule
\multirow{10}{*}{\SI{10}{\percent}} & Adult Income & $0.496$ & $0.497$ & $0.498$ & $0.498$ & $0.499$ \\
 & Diabetes 130-US & $0.507$ & $0.507$ & $0.508$ & $0.508$ & $0.507$ \\
 & Breast Cancer & $0.518$ & $0.511$ & $0.518$ & $0.516$ & $0.516$ \\
 & Heart Disease & $\mathbf{0.421}$ & $\mathbf{0.387}$ & $\mathbf{0.438}$ & $\mathbf{0.429}$ & $\mathbf{0.417}$ \\
 & German Credit & $0.494$ & $0.489$ & $0.482$ & $0.481$ & $0.473$ \\
 & Bank Marketing & $0.503$ & $0.505$ & $0.504$ & $0.504$ & $0.504$ \\
 & Wine Quality & $0.458$ & $0.456$ & $0.465$ & $0.465$ & $0.458$ \\
 & Phoneme & $0.516$ & $0.513$ & $0.519$ & $0.518$ & $0.518$ \\
 & Magic Telescope & $0.498$ & $0.497$ & $0.502$ & $0.502$ & $0.501$ \\
 & Electricity & $0.497$ & $0.500$ & $0.501$ & $0.501$ & $0.500$ \\
\bottomrule
\end{tabular}
\end{table}

\subsection{Sensitivity and calibration of $\Mtwo$}
\label{app:m2_baseline}
Table~\ref{tab:m2_baseline_sensitivity} reports $\Mtwo$ under the median (used in this paper) and arithmetic mean baselines for the retain set. Under the median baseline, all $12$ method-fraction conditions are significant at $p<0.001$ on the pooled Wilcoxon signed-rank test. Under the mean baseline, only $3$ of $12$ remain significant and $5$ of $12$ reverse sign. 

The retain distribution places the mean roughly $0.0023$ below the median for NegGrad$+$ and $0.0013$ below for Fine-Tuning at $\ff=\SI{1}{\percent}$. This displacement is comparable to the gap itself, lowering the calibration baseline and masking the residual signal that the median baseline detects with large effect sizes.

\begin{table}[H]
\footnotesize
\centering
\caption{Sensitivity of $\Mtwo$ to retain baseline choice across all forget fractions ($N=100$ per condition; Wilcoxon signed-rank). Under the median baseline (used in this paper), all $12$ conditions are significant at $p<0.001$. Under the mean baseline, only $3$ of $12$ remain significant ($^*p<0.05$, $^{**}p<0.01$, $^{***}p<0.001$) and $5$ of $12$ reverse sign.}
\label{tab:m2_baseline_sensitivity}
\setlength{\tabcolsep}{7pt}
\begin{tabular}{llrr rrr}
\toprule
 & & \multicolumn{2}{c}{Median baseline} & \multicolumn{3}{c}{Mean baseline} \\
\cmidrule(lr){3-4}\cmidrule(lr){5-7}
Method & $\ff$ & $\Mtwo$ & $\rrb$ & $\Mtwo$ & $p$ & $\rrb$ \\
\midrule
Gradient Ascent & \SI{1}{\percent}  & $-0.00077$ & $+0.51$ & $+0.00040$ & $0.270$ & $+0.13$ \\
NegGrad$+$      & \SI{1}{\percent}  & $-0.00470$ & $+0.87$ & $-0.00242$ & $<0.001^{***}$ & $+0.43$ \\
Fine-Tuning     & \SI{1}{\percent}  & $-0.00112$ & $+0.71$ & $+0.00013$ & $0.477$ & $+0.08$ \\
SCRUB           & \SI{1}{\percent}  & $-0.00121$ & $+0.71$ & $-0.00001$ & $0.633$ & $+0.06$ \\
\addlinespace
Gradient Ascent & \SI{5}{\percent}  & $-0.00097$ & $+0.54$ & $+0.00053$ & $0.074$ & $+0.21$ \\
NegGrad$+$      & \SI{5}{\percent}  & $-0.00203$ & $+0.76$ & $+0.00033$ & $0.765$ & $+0.03$ \\
Fine-Tuning     & \SI{5}{\percent}  & $-0.00134$ & $+0.73$ & $-0.00003$ & $0.765$ & $+0.03$ \\
SCRUB           & \SI{5}{\percent}  & $-0.00122$ & $+0.71$ & $+0.00007$ & $0.929$ & $+0.01$ \\
\addlinespace
Gradient Ascent & \SI{10}{\percent} & $-0.00167$ & $+0.90$ & $-0.00017$ & $0.199$ & $+0.15$ \\
NegGrad$+$      & \SI{10}{\percent} & $-0.00171$ & $+0.96$ & $-0.00006$ & $0.698$ & $+0.04$ \\
Fine-Tuning     & \SI{10}{\percent} & $-0.00210$ & $+0.93$ & $-0.00064$ & $0.010^{*}$  & $+0.30$ \\
SCRUB           & \SI{10}{\percent} & $-0.00207$ & $+0.94$ & $-0.00061$ & $0.008^{**}$ & $+0.30$ \\
\bottomrule
\end{tabular}
\end{table}

\begin{figure}[H]
  \centering
  \includegraphics[width=\linewidth]{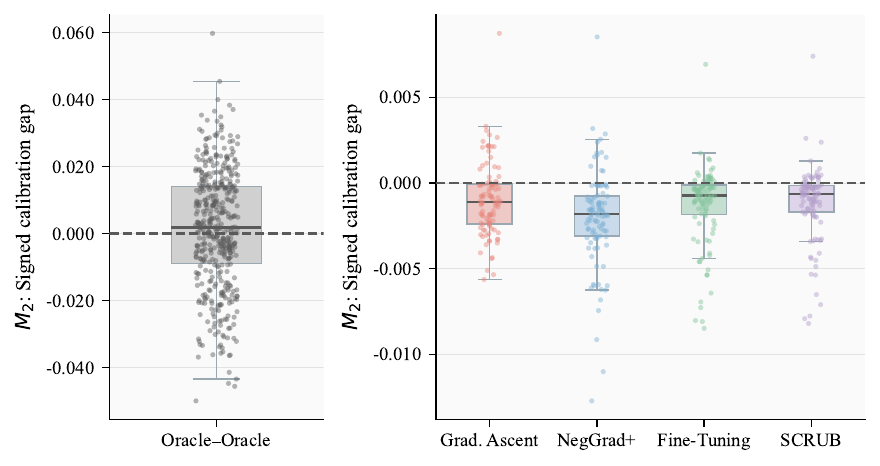}
  \caption{Empirical calibration of $\Mtwo$ at $\ff=\SI{5}{\percent}$. Dashed line: null of zero. Left: $\Mtwo$ across all $\binom{10}{2}=45$ oracle pairs per dataset, pooled across 10 datasets (450 values); approximately centred on zero, confirming the metric is well-calibrated under correct retraining. Right: $\Mtwo$ for the four unlearning methods ($N=100$ per method; 10 datasets $\times$ 10 seeds) on a finer y-axis than Fig.~\ref{fig:fig0}; all four produce consistently negative gaps. Boxes: IQR; whiskers: $1.5\times$ IQR.}
  \label{fig:m2_calibration}
\end{figure}

\paragraph{Empirical calibration of the null.}
The null hypothesis $\Mtwo=0$ (Section~\ref{subsec:lens1}) is also empirically achievable. Applying $\Mtwo$ to pairs of independently retrained oracles (no unlearning applied to either model) yields a distribution approximately centred on zero (Fig.~\ref{fig:m2_calibration}, left), confirming that the metric does not deviate from zero by construction. The right panel shows the four unlearning methods at $\ff=\SI{5}{\percent}$ on a finer y-axis than Fig.~\ref{fig:fig0}, showing the within-method distribution alongside the calibrated null: all four sit below zero.

\subsection{Robustness to mini-batch training}
\label{app:minibatch}
The main experiments use full-batch gradient descent to isolate unlearning dynamics from optimisation noise. To assess robustness to stochastic optimisation, we repeated the $\ff=\SI{5}{\percent}$ experiment using mini-batch SGD (batch size 128, 5 seeds per dataset, $N=50$) and applied the same LMM specification used in the main analysis (Section~\ref{sec:stats}). 

For $\Mtwo$, the random-effect variance was not identifiable for Gradient Ascent, Fine-Tuning, and SCRUB (singular REML fit) because within-dataset clustering was close to perfect at this sample size; we therefore report the Wilcoxon signed-rank test as a fallback for these three methods and the LMM result for NegGrad$+$. Under mini-batch training, $\Mtwo$ reverses sign for three of four methods (Table~\ref{tab:minibatch}), confirming that the paired-seed calibration baseline is invalidated when batch ordering introduces stochastic variation between the original model and oracle. NegGrad$+$ is the only method for which $\Mtwo$ remains negative, but its magnitude is small ($-0.00195$, $\rrb=+0.02$) and does not differ significantly from zero under either test. 

$\Mfour$ remains above $0.50$ for all four methods (range $0.519$ to $0.538$). Under the LMM, NegGrad$+$ is significant at $p<0.001$ (ICC $=0.95$); the other three methods reach $p\leq 0.020$ under the Wilcoxon test but only marginal significance ($p$ between $0.072$ and $0.089$) under the LMM, where very high within-dataset clustering (ICC $\geq 0.97$) absorbs most of the between-seed signal. The residual signal on $\Mfour$ therefore persists under stochastic optimisation, though the LMM is conservative at this sample size.

\begin{table}[H]\footnotesize
\centering
\caption{$\Mtwo$ and $\Mfour$ under mini-batch training at $\ff=\SI{5}{\percent}$ (batch size 128; $N=50$ per method; 10 datasets~$\times$~5 seeds). $p$-values from a linear mixed-effects model with random intercept for dataset; Wilcoxon signed-rank $p$-values reported as a fallback, and for $\Mtwo$ replace the LMM $p$-value for Gradient Ascent, Fine-Tuning, and SCRUB where the random-effect variance was unidentifiable (singular REML fit). Significance: $^{***}p<0.001$.}
\label{tab:minibatch}
\setlength{\tabcolsep}{8pt}
\begin{tabular}{lrrrr}
\toprule
Method & gap/rank & $p_{\mathrm{LMM}}$ & $p_{\mathrm{Wilc}}$ & $\rrb$ \\
\midrule
\multicolumn{5}{l}{\textbf{$\Mtwo$ (signed calibration gap, null $=0$)}} \\
Gradient Ascent & $+0.00575$ & n/a    & $0.003$ & $+0.48$ \\
NegGrad$+$      & $-0.00195$ & $0.702$ & $0.893$ & $+0.02$ \\
Fine-Tuning     & $+0.00536$ & n/a    & $0.002$ & $+0.49$ \\
SCRUB           & $+0.00640$ & n/a    & $0.001$ & $+0.51$ \\
\addlinespace
\multicolumn{5}{l}{\textbf{$\Mfour$ (percentile rank, null $=0.50$)}} \\
Gradient Ascent & $0.522$ & $0.072$         & $0.010$         & $+0.42$ \\
NegGrad$+$      & $0.538$ & $<0.001^{***}$  & $<0.001^{***}$  & $+1.00$ \\
Fine-Tuning     & $0.522$ & $0.075$         & $0.010$         & $+0.41$ \\
SCRUB           & $0.519$ & $0.089$         & $0.020$         & $+0.38$ \\
\bottomrule
\end{tabular}
\end{table}

\subsection{Hyperparameter sensitivity}
\label{app:sensitivity}
To assess robustness to the unlearning learning rate, we repeated the primary $\ff=\SI{5}{\percent}$ experiment with $\eta_u\in\{10^{-4},5\times 10^{-4},10^{-3}\}$, scaling epoch counts proportionally so that the total update magnitude is comparable across configurations (Table~\ref{tab:sensitivity}). $\Mtwo$ is negative for 7 of 10 datasets across all configurations; Heart Disease and German Credit show occasional positive values consistent with the exceptions identified in Section~\ref{subsec:residuals}, and Electricity produces a single positive value at the most aggressive learning rate. 
$\Mfour$ follows the same per-dataset pattern as the main analysis (Fig.~\ref{fig:per_dataset_m4}). Datasets above $0.50$ there remain elevated across learning rates, and datasets near or below it remain near the null, with the only boundary crossings two Gradient Ascent cells within $0.012$ of $0.50$ at the most aggressive learning rate.
The combined criterion ($\Mtwo<0$ and $\Mfour>0.50$) holds for the same datasets in most configurations, indicating that the residual signal is not specific to the learning rate used in the main experiments.

\begin{table}[H]
\centering
\rotatebox{90}{%
\begin{minipage}{\textheight}
\caption{Hyperparameter sensitivity at $\ff=\SI{5}{\percent}$. $\Mtwo$ signed calibration and $\Mfour$ percentile rank across three learning rates ($\eta_u\in\{10^{-4},5\times 10^{-4},10^{-3}\}$) with epoch counts scaled proportionally. $\Mtwo$ is negative for 7 of 10 datasets across all configurations; $\Mfour$ follows the same dataset-level pattern as the main analysis. Holds = number of configurations (out of 3) in which both $\Mtwo<0$ and $\Mfour>0.50$; $\checkmark$ marks rows where the criterion holds at all three learning rates. (Part 1 of 2; continued in Table~\ref{tab:sensitivity_p2}.)}
\label{tab:sensitivity}
\centering
\setlength{\tabcolsep}{6pt}
\begin{tabular}{llccccccc}
\toprule
Dataset & Method & \multicolumn{3}{c}{$\Mtwo$ gap} & \multicolumn{3}{c}{$\Mfour$ rank} & Holds \\
\cmidrule(lr){3-5}\cmidrule(lr){6-8}
& & $\eta=10^{-4}$ & $\eta=5{\times}10^{-4}$ & $\eta=10^{-3}$ & $\eta=10^{-4}$ & $\eta=5{\times}10^{-4}$ & $\eta=10^{-3}$ & \\
\midrule
Adult Income & Gradient Ascent & $-0.0001$ & $-0.0024$ & $-0.0211$ & $0.488$ & $0.489$ & $0.498$ & 0/3 \\
 & NegGrad$+$ & $-0.0001$ & $-0.0019$ & $-0.0072$ & $0.488$ & $0.489$ & $0.492$ & 0/3 \\
 & Fine-Tuning & $-0.0001$ & $-0.0005$ & $-0.0050$ & $0.488$ & $0.486$ & $0.485$ & 0/3 \\
 & SCRUB & $-0.0001$ & $-0.0005$ & $-0.0048$ & $0.488$ & $0.486$ & $0.484$ & 0/3 \\
\midrule
Diabetes 130-US & Gradient Ascent & $-0.0000$ & $-0.0001$ & $-0.0029$ & $0.502$ & $0.502$ & $0.502$ & $\checkmark$ 3/3 \\
 & NegGrad$+$ & $-0.0000$ & $-0.0002$ & $-0.0026$ & $0.502$ & $0.503$ & $0.506$ & $\checkmark$ 3/3 \\
 & Fine-Tuning & $-0.0000$ & $-0.0003$ & $-0.0026$ & $0.502$ & $0.504$ & $0.505$ & $\checkmark$ 3/3 \\
 & SCRUB & $-0.0000$ & $-0.0003$ & $-0.0023$ & $0.502$ & $0.504$ & $0.505$ & $\checkmark$ 3/3 \\
\midrule
Breast Cancer & Gradient Ascent & $-0.0004$ & $-0.0016$ & $-0.0484$ & $0.610$ & $0.602$ & $0.559$ & $\checkmark$ 3/3 \\
 & NegGrad$+$ & $-0.0004$ & $-0.0036$ & $-0.0522$ & $0.610$ & $0.600$ & $0.597$ & $\checkmark$ 3/3 \\
 & Fine-Tuning & $-0.0005$ & $-0.0005$ & $-0.0009$ & $0.611$ & $0.610$ & $0.602$ & $\checkmark$ 3/3 \\
 & SCRUB & $-0.0005$ & $-0.0007$ & $-0.0088$ & $0.611$ & $0.609$ & $0.609$ & $\checkmark$ 3/3 \\
\midrule
Heart Disease & Gradient Ascent & $-0.0023$ & $+0.0012$ & $-0.0226$ & $0.553$ & $0.542$ & $0.531$ & 2/3 \\
 & NegGrad$+$ & $-0.0019$ & $+0.0000$ & $-0.0851$ & $0.552$ & $0.540$ & $0.593$ & 2/3 \\
 & Fine-Tuning & $-0.0028$ & $-0.0021$ & $+0.0011$ & $0.555$ & $0.559$ & $0.568$ & 2/3 \\
 & SCRUB & $-0.0026$ & $-0.0018$ & $-0.0065$ & $0.554$ & $0.554$ & $0.562$ & $\checkmark$ 3/3 \\
\midrule
German Credit & Gradient Ascent & $-0.0005$ & $+0.0019$ & $-0.0159$ & $0.547$ & $0.537$ & $0.490$ & 1/3 \\
 & NegGrad$+$ & $-0.0003$ & $+0.0020$ & $-0.0560$ & $0.547$ & $0.531$ & $0.540$ & 2/3 \\
 & Fine-Tuning & $-0.0007$ & $+0.0011$ & $+0.0045$ & $0.548$ & $0.543$ & $0.540$ & 1/3 \\
 & SCRUB & $-0.0008$ & $+0.0009$ & $+0.0038$ & $0.548$ & $0.544$ & $0.550$ & 1/3 \\
\bottomrule
\end{tabular}
\end{minipage}}
\end{table}

\begin{table}[H]
\centering
\rotatebox{90}{%
\begin{minipage}{\textheight}
\centering
\caption{Hyperparameter sensitivity, continued from Table~\ref{tab:sensitivity}. Holds column as defined in Table~\ref{tab:sensitivity}.}
\label{tab:sensitivity_p2}
\centering
\setlength{\tabcolsep}{4pt}
\renewcommand{\arraystretch}{0.95}
\begin{tabular}{llccccccc}
\toprule
Dataset & Method & \multicolumn{3}{c}{$\Mtwo$ gap} & \multicolumn{3}{c}{$\Mfour$ rank} & Holds \\
\cmidrule(lr){3-5}\cmidrule(lr){6-8}
& & $\eta=10^{-4}$ & $\eta=5{\times}10^{-4}$ & $\eta=10^{-3}$ & $\eta=10^{-4}$ & $\eta=5{\times}10^{-4}$ & $\eta=10^{-3}$ & \\
\midrule
Bank Marketing & Gradient Ascent & $-0.0001$ & $-0.0034$ & $-0.0307$ & $0.495$ & $0.494$ & $0.495$ & 0/3 \\
 & NegGrad$+$ & $-0.0001$ & $-0.0025$ & $-0.0164$ & $0.495$ & $0.495$ & $0.496$ & 0/3 \\
 & Fine-Tuning & $-0.0003$ & $-0.0063$ & $-0.0247$ & $0.495$ & $0.494$ & $0.492$ & 0/3 \\
 & SCRUB & $-0.0003$ & $-0.0061$ & $-0.0196$ & $0.495$ & $0.494$ & $0.493$ & 0/3 \\
\midrule
Wine Quality & Gradient Ascent & $-0.0025$ & $-0.0024$ & $-0.0123$ & $0.506$ & $0.504$ & $0.504$ & $\checkmark$ 3/3 \\
 & NegGrad$+$ & $-0.0024$ & $-0.0042$ & $-0.0252$ & $0.505$ & $0.499$ & $0.501$ & 2/3 \\
 & Fine-Tuning & $-0.0025$ & $-0.0025$ & $-0.0034$ & $0.507$ & $0.508$ & $0.507$ & $\checkmark$ 3/3 \\
 & SCRUB & $-0.0026$ & $-0.0025$ & $-0.0033$ & $0.505$ & $0.504$ & $0.507$ & $\checkmark$ 3/3 \\
\midrule
Phoneme & Gradient Ascent & $-0.0003$ & $-0.0020$ & $-0.0618$ & $0.496$ & $0.497$ & $0.498$ & 0/3 \\
 & NegGrad$+$ & $-0.0003$ & $-0.0074$ & $-0.0216$ & $0.496$ & $0.497$ & $0.498$ & 0/3 \\
 & Fine-Tuning & $-0.0003$ & $-0.0013$ & $-0.0068$ & $0.496$ & $0.497$ & $0.498$ & 0/3 \\
 & SCRUB & $-0.0003$ & $-0.0011$ & $-0.0062$ & $0.496$ & $0.496$ & $0.498$ & 0/3 \\
\midrule
Magic Telescope & Gradient Ascent & $-0.0004$ & $-0.0017$ & $-0.0368$ & $0.486$ & $0.486$ & $0.490$ & 0/3 \\
 & NegGrad$+$ & $-0.0005$ & $-0.0017$ & $-0.0094$ & $0.487$ & $0.487$ & $0.486$ & 0/3 \\
 & Fine-Tuning & $-0.0004$ & $-0.0011$ & $-0.0062$ & $0.486$ & $0.486$ & $0.484$ & 0/3 \\
 & SCRUB & $-0.0004$ & $-0.0011$ & $-0.0061$ & $0.486$ & $0.486$ & $0.484$ & 0/3 \\
\midrule
Electricity & Gradient Ascent & $-0.0001$ & $-0.0006$ & $+0.0078$ & $0.496$ & $0.498$ & $0.502$ & 0/3 \\
 & NegGrad$+$ & $-0.0001$ & $-0.0011$ & $-0.0024$ & $0.497$ & $0.497$ & $0.495$ & 0/3 \\
 & Fine-Tuning & $-0.0001$ & $-0.0002$ & $-0.0011$ & $0.496$ & $0.497$ & $0.498$ & 0/3 \\
 & SCRUB & $-0.0001$ & $-0.0002$ & $-0.0009$ & $0.496$ & $0.497$ & $0.498$ & 0/3 \\
\bottomrule
\end{tabular}
\end{minipage}}
\end{table}

\subsection{Robustness to forget-set sampling}
\label{app:forget_seed}

The main experiments use a single forget set per (dataset, $\ff$) condition. To check that the reported results do not depend on this choice, we repeated the pipeline on all ten datasets and all three forget fractions using five additional forget-set seeds (999 to 1003), training a fresh oracle for each ($N=3{,}000$ runs; Table~\ref{tab:forget_seed_sensitivity}). $\Mtwo$ is negative across all 20 (method, forget-set seed) combinations in 21 of 30 cells. The 9 cells with at least one positive value involve the same exceptions identified in the main analysis (Heart Disease, German Credit, Electricity); Wine Quality and Adult Income additionally show occasional positive values under this independent sampling. The across-seed standard deviation (median across methods) is largest on the smallest forget sets, reaching $0.0048$ for $\Mtwo$ and $0.112$ for $\Mfour$ at $|\Df|\leq 12$, and falls to at most $0.0007$ for $\Mtwo$ and $0.015$ for $\Mfour$ once $|\Df|\geq 1{,}000$. Per-cell $\Mfour$ can therefore be noisy at small $|\Df|$, including the forget-set sizes that arise from individual erasure requests under GDPR Article~17, but the population-level inference in Table~\ref{tab:cross_ff} is unaffected because the LMM averages over this variation.

\begin{table}[H]
\centering
\caption{Variability of $\Mtwo$ and $\Mfour$ across five forget-set seeds on all ten datasets. Each row aggregates 20 values (4 methods $\times$ 5 forget-set seeds), each averaged over 5 training seeds ($N=100$ runs per row). $|\Df|$: forget-set size. sign: number of the 20 values that are negative. $\Mtwo$\,SD, $\Mfour$\,SD: median across the four methods of the across-seed standard deviation. Cells below $20/20$ are bold.}
\label{tab:forget_seed_sensitivity}
\setlength{\tabcolsep}{6pt}
\begin{tabular}{l r r c c c}
\toprule
Dataset & $\ff$ & $|\Df|$ & sign & $\Mtwo$\,SD & $\Mfour$\,SD \\
\midrule
\multirow{3}{*}{Heart Disease}   & \SI{1}{\percent}  & 10   & \textbf{19/20} & $0.0048$ & $0.088$ \\
                                  & \SI{5}{\percent}  & 12   & 20/20          & $0.0035$ & $0.070$ \\
                                  & \SI{10}{\percent} & 24   & 20/20          & $0.0022$ & $0.057$ \\
\addlinespace
\multirow{3}{*}{Breast Cancer}   & \SI{1}{\percent}  & 10   & 20/20          & $0.0011$ & $0.018$ \\
                                  & \SI{5}{\percent}  & 22   & 20/20          & $0.0005$ & $0.048$ \\
                                  & \SI{10}{\percent} & 45   & 20/20          & $0.0005$ & $0.020$ \\
\addlinespace
\multirow{3}{*}{German Credit}   & \SI{1}{\percent}  & 10   & \textbf{18/20} & $0.0041$ & $0.062$ \\
                                  & \SI{5}{\percent}  & 40   & \textbf{16/20} & $0.0029$ & $0.061$ \\
                                  & \SI{10}{\percent} & 80   & \textbf{19/20} & $0.0012$ & $0.024$ \\
\addlinespace
\multirow{3}{*}{Wine Quality}    & \SI{1}{\percent}  & 12   & \textbf{17/20} & $0.0031$ & $0.112$ \\
                                  & \SI{5}{\percent}  & 63   & \textbf{17/20} & $0.0019$ & $0.039$ \\
                                  & \SI{10}{\percent} & 127  & 20/20          & $0.0015$ & $0.015$ \\
\addlinespace
\multirow{3}{*}{Phoneme}         & \SI{1}{\percent}  & 43   & 20/20          & $0.0015$ & $0.045$ \\
                                  & \SI{5}{\percent}  & 216  & 20/20          & $0.0007$ & $0.014$ \\
                                  & \SI{10}{\percent} & 432  & 20/20          & $0.0004$ & $0.014$ \\
\addlinespace
\multirow{3}{*}{Magic Telescope} & \SI{1}{\percent}  & 152  & 20/20          & $0.0013$ & $0.020$ \\
                                  & \SI{5}{\percent}  & 760  & 20/20          & $0.0005$ & $0.016$ \\
                                  & \SI{10}{\percent} & 1521 & 20/20          & $0.0007$ & $0.015$ \\
\addlinespace
\multirow{3}{*}{Bank Marketing}  & \SI{1}{\percent}  & 361  & 20/20          & $0.0008$ & $0.013$ \\
                                  & \SI{5}{\percent}  & 1808 & 20/20          & $0.0007$ & $0.005$ \\
                                  & \SI{10}{\percent} & 3616 & 20/20          & $0.0006$ & $0.012$ \\
\addlinespace
\multirow{3}{*}{Electricity}     & \SI{1}{\percent}  & 362  & 20/20          & $0.0003$ & $0.021$ \\
                                  & \SI{5}{\percent}  & 1812 & \textbf{15/20} & $0.0005$ & $0.007$ \\
                                  & \SI{10}{\percent} & 3624 & \textbf{17/20} & $0.0004$ & $0.009$ \\
\addlinespace
\multirow{3}{*}{Adult Income}    & \SI{1}{\percent}  & 390  & \textbf{18/20} & $0.0008$ & $0.026$ \\
                                  & \SI{5}{\percent}  & 1953 & 20/20          & $0.0003$ & $0.012$ \\
                                  & \SI{10}{\percent} & 3907 & 20/20          & $0.0005$ & $0.008$ \\
\addlinespace
\multirow{3}{*}{Diabetes 130-US} & \SI{1}{\percent}  & 814  & 20/20          & $0.0000$ & $0.016$ \\
                                  & \SI{5}{\percent}  & 4070 & 20/20          & $0.0000$ & $0.009$ \\
                                  & \SI{10}{\percent} & 8141 & 20/20          & $0.0000$ & $0.009$ \\
\bottomrule
\end{tabular}
\end{table}

\subsection{Per-dataset representation shift}
\label{app:per_dataset_shift}
\begin{figure}[H]
  \centering
  \includegraphics[width=0.95\linewidth]{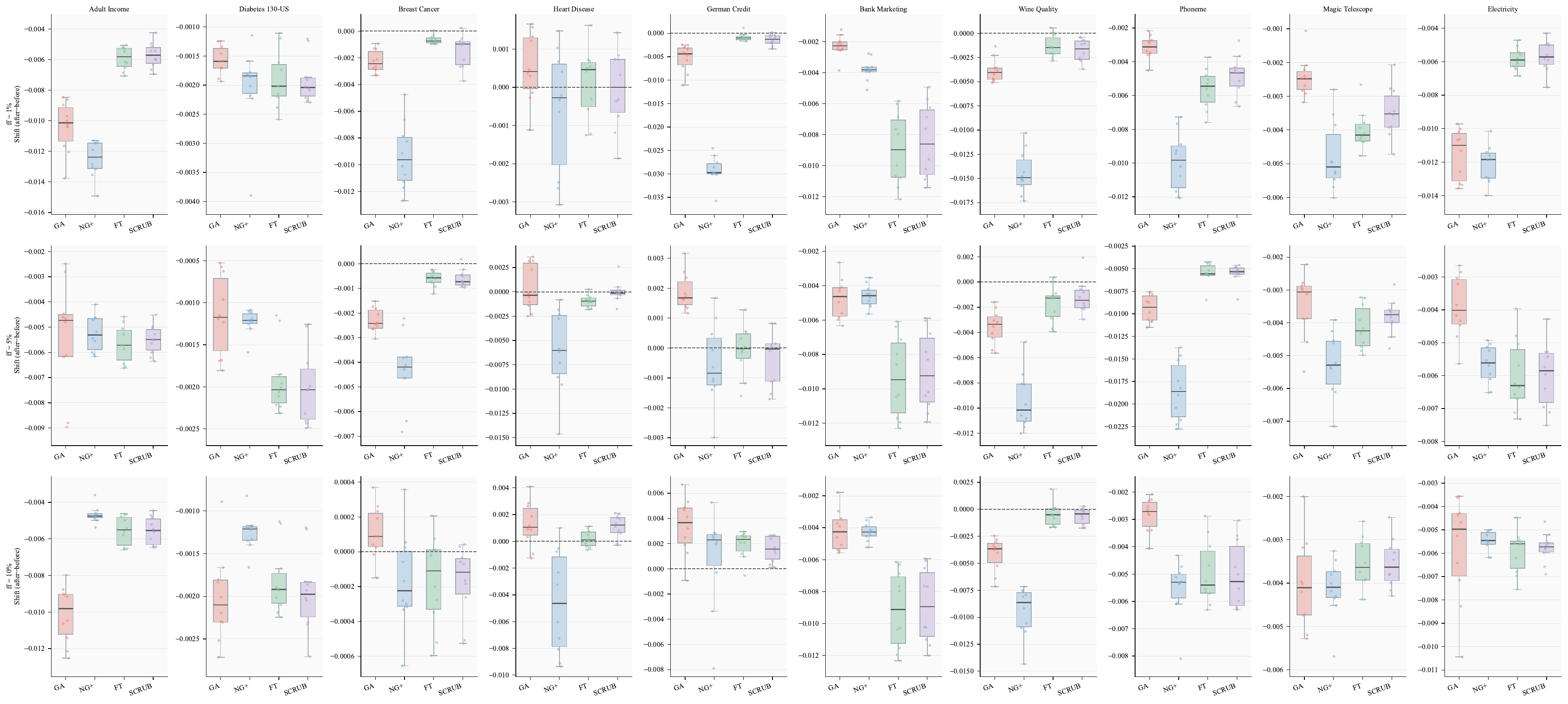}
  \caption{Per-dataset representation shift ($\Mthree$) across all three forget fractions (rows: $\ff=\SI{1}{\percent},\SI{5}{\percent},\SI{10}{\percent}$; columns: 10 datasets; 10 training seeds per condition). Shift = mean change in cosine similarity to the oracle on forget-set records, after minus before unlearning. Dashed line: zero shift.}
  \label{fig:per_dataset_shift}
\end{figure}

Fig.~\ref{fig:per_dataset_shift} shows the per-dataset representation shift $\Mthree$ across all three forget fractions and ten datasets. The negative direction dominates across datasets and forget fractions; Heart Disease and German Credit show the anomalous patterns discussed in Section~\ref{subsec:residuals}.

\subsection{Per-dataset $\Mfour$ distribution}
\label{app:per_dataset_m4}
\begin{figure}[H]
  \centering
  \includegraphics[width=\linewidth]{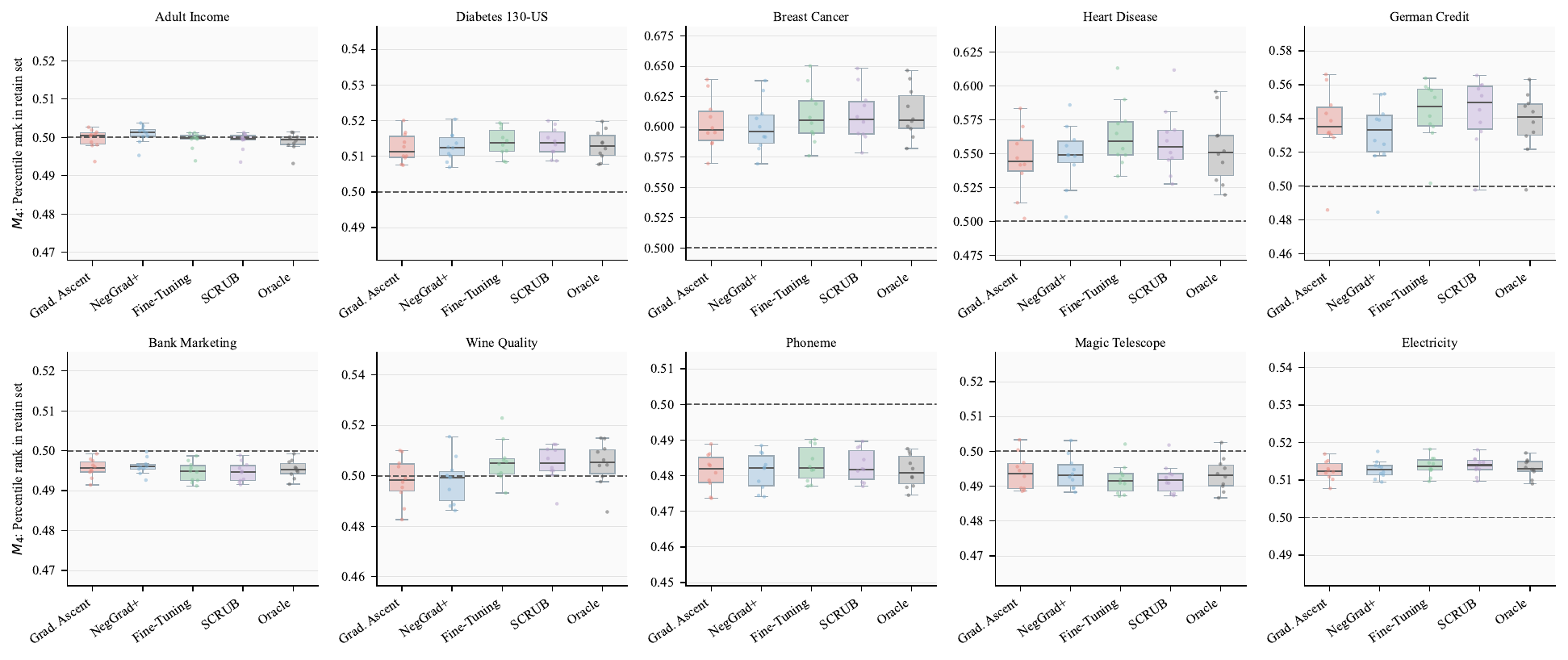}
  \caption{Per-dataset oracle-free percentile rank ($\Mfour$) at $\ff=\SI{5}{\percent}$ (10 training seeds per dataset; the oracle baseline is the fifth box per panel for direct comparison). Dashed line: $0.50$ reference.}
  \label{fig:per_dataset_m4}
\end{figure}

Fig.~\ref{fig:per_dataset_m4} shows the per-dataset distribution of $\Mfour$ at $\ff=\SI{5}{\percent}$. The four unlearning methods produce similar distributions to the oracle baseline within each dataset, with deviations from $0.50$ driven by the dataset's feature-space geometry rather than the unlearning method.

\subsection{Comparison with Bad Teacher}
\label{app:bad_teacher}
Gradient Ascent, NegGrad$+$, Fine-Tuning, and SCRUB all modify the original model's parameters directly. Bad Teacher~\cite{chundawat2023can} operates differently, training a student to match a randomly initialised teacher on forget-set records. Testing whether the same residuals appear under this mechanism indicates whether the discordance is method-specific or structural.

\paragraph{Implementation.}
The implementation follows Chundawat et al.~\cite{chundawat2023can}. A randomly initialised \texttt{TabularMLP} is frozen as the teacher. A student copy of the original model is trained for 10 epochs using Adam at $\eta_u=5\times 10^{-4}$, $\alpha=0.6$, $T=2.0$, matching the SCRUB hyperparameters. On forget-set records the student is trained to match the random teacher's output distribution; on retain-set records it minimises both the KL divergence to the random teacher and the cross-entropy loss on true labels. The original model provides only the student's initialisation, and no competent teacher is used: this adapts the incompetent-teacher idea of~\cite{chundawat2023can} to a single-teacher objective in which the retain set is preserved by the true labels rather than by a competent teacher. A single unlearning seed ($j=100$) is used, matching the design of the main experiment. To assess sensitivity to teacher initialisation, we repeated the analysis with two additional teacher seeds (101, 102). $\Mtwo$ remained negative for all ten datasets at all three forget fractions across all three seeds. $\Mfour$ values were stable across seeds, with maximum within-dataset variation below $0.007$. These results confirm that the findings are not specific to the choice of teacher initialisation.

\paragraph{Results.}
Bad Teacher passes output-level evaluation at all three forget fractions. $\Mtwo$ detects significant residuals throughout ($p<0.01$; Table~\ref{tab:bt_repr}), meaning forget-set records remain displaced from where the oracle places them. $\Mfour$ does not reach significance at any fraction, with mean ranks remaining above $0.50$ ($0.528$, $0.514$, $0.505$ at $\ff=\SI{1}{\percent}$, \SI{5}{\percent}, \SI{10}{\percent}), consistent with the attenuation observed in Section~\ref{subsec:metric_divergence}. Paired comparisons on $\Mfour$ (Table~\ref{tab:bt_paired}) show that Bad Teacher displaces forget-set records further from the retain set than all four approximate methods at $\ff=\SI{1}{\percent}$ (all $p\leq0.01$). At larger fractions no comparison reaches significance. This stronger displacement does not achieve complete erasure: $\Mtwo$ remains negative at every fraction, confirming forget-set records still fall below the retain-set baseline in the oracle's representation space. The discordance therefore holds across five methods spanning direct parameter modification and distillation-based unlearning, indicating that current approximate methods are optimised for output-level criteria that leave intermediate-layer representations unconstrained.

\begin{table}[H]
\centering
\footnotesize
\caption{Dataset-level $\Mtwo$ and $\Mfour$ statistics for approximate unlearning methods and Bad Teacher. One-sample Wilcoxon signed-rank test on $N=10$ dataset-level means, averaged over 10 training seeds per dataset ($\Mtwo$ null $=0$; $\Mfour$ null $=0.50$).}
\label{tab:bt_repr}
\setlength{\tabcolsep}{7pt}
\begin{tabular}{lcccccc}
\toprule
 & \multicolumn{3}{c}{$\Mtwo$} & \multicolumn{3}{c}{$\Mfour$} \\
\cmidrule(lr){2-4}\cmidrule(lr){5-7}
Method & mean & $p$ & $\rrb$ & mean & $p$ & $\rrb$ \\
\midrule
\multicolumn{7}{l}{\textbf{$\ff=\SI{1}{\percent}$}} \\
Grad.\ Ascent & $-0.00077$ & $0.160$ & $+0.527$ & $0.5328$ & $0.275$ & $+0.418$ \\
NegGrad$+$    & $-0.00470$ & $0.014$ & $+0.855$ & $0.5324$ & $0.322$ & $+0.382$ \\
Fine-Tuning   & $-0.00112$ & $0.010$ & $+0.891$ & $0.5383$ & $0.193$ & $+0.491$ \\
SCRUB         & $-0.00121$ & $0.014$ & $+0.855$ & $0.5376$ & $0.193$ & $+0.491$ \\
Bad Teacher   & $-0.00313$ & $0.002$ & $+1.000$ & $0.5277$ & $0.322$ & $+0.382$ \\
\addlinespace
\multicolumn{7}{l}{\textbf{$\ff=\SI{5}{\percent}$}} \\
Grad.\ Ascent & $-0.00097$ & $0.131$ & $+0.564$ & $0.5178$ & $0.322$ & $+0.382$ \\
NegGrad$+$    & $-0.00203$ & $0.037$ & $+0.745$ & $0.5175$ & $0.275$ & $+0.418$ \\
Fine-Tuning   & $-0.00134$ & $0.027$ & $+0.782$ & $0.5220$ & $0.193$ & $+0.491$ \\
SCRUB         & $-0.00122$ & $0.027$ & $+0.782$ & $0.5214$ & $0.232$ & $+0.455$ \\
Bad Teacher   & $-0.00370$ & $0.002$ & $+1.000$ & $0.5142$ & $0.275$ & $+0.418$ \\
\addlinespace
\multicolumn{7}{l}{\textbf{$\ff=\SI{10}{\percent}$}} \\
Grad.\ Ascent & $-0.00167$ & $0.002$ & $+1.000$ & $0.5058$ & $0.557$ & $+0.236$ \\
NegGrad$+$    & $-0.00171$ & $0.002$ & $+1.000$ & $0.5058$ & $0.557$ & $+0.236$ \\
Fine-Tuning   & $-0.00210$ & $0.002$ & $+1.000$ & $0.5062$ & $0.557$ & $+0.236$ \\
SCRUB         & $-0.00207$ & $0.002$ & $+1.000$ & $0.5059$ & $0.695$ & $+0.164$ \\
Bad Teacher   & $-0.00480$ & $0.002$ & $+1.000$ & $0.5046$ & $0.922$ & $+0.055$ \\
\bottomrule
\end{tabular}
\end{table}

\begin{table}[H]
\centering
\footnotesize
\caption{Paired $\Mfour$ comparisons between Bad Teacher and each approximate method. $\Delta$: Bad Teacher mean minus method mean. Paired Wilcoxon signed-rank test on $N=10$ dataset-level means. Negative $\Delta$ indicates less residual integration for Bad Teacher. Significance: $^{**}p\leq0.01$; ns $p\geq0.05$.}
\label{tab:bt_paired}
\setlength{\tabcolsep}{7pt}
\begin{tabular}{c l r r r l}
\toprule
$\ff$ & Method & $\Mfour$ mean & $\Delta$ & $p$ & sig \\
\midrule
\multirow{4}{*}{\SI{1}{\percent}} & Grad.\ Ascent & $0.5328$ & $-0.0051$ & $0.010$ & $**$ \\
 & NegGrad$+$    & $0.5324$ & $-0.0048$ & $0.006$ & $**$ \\
 & Fine-Tuning   & $0.5383$ & $-0.0106$ & $0.006$ & $**$ \\
 & SCRUB         & $0.5376$ & $-0.0099$ & $0.010$ & $**$ \\
\midrule
\multirow{4}{*}{\SI{5}{\percent}} & Grad.\ Ascent & $0.5178$ & $-0.0036$ & $0.695$ & ns \\
 & NegGrad$+$    & $0.5175$ & $-0.0033$ & $0.695$ & ns \\
 & Fine-Tuning   & $0.5220$ & $-0.0077$ & $0.322$ & ns \\
 & SCRUB         & $0.5214$ & $-0.0071$ & $0.375$ & ns \\
\midrule
\multirow{4}{*}{\SI{10}{\percent}} & Grad.\ Ascent & $0.5058$ & $-0.0013$ & $0.557$ & ns \\
 & NegGrad$+$    & $0.5058$ & $-0.0013$ & $0.922$ & ns \\
 & Fine-Tuning   & $0.5062$ & $-0.0017$ & $0.084$ & ns \\
 & SCRUB         & $0.5059$ & $-0.0014$ & $0.557$ & ns \\
\bottomrule
\end{tabular}
\end{table}

\subsection{$\Mfour$ on large-scale tabular datasets}
\label{app:largescale}
Table~\ref{tab:largescale} reports $\Mfour$ on Covertype and Higgs (100{,}000 samples each, $\ff=\SI{5}{\percent}$, batch size 256, 5 seeds).

\begin{table}[H]
\centering
\caption{$\Mfour$ on large-scale datasets (100{,}000 samples, $\ff=\SI{5}{\percent}$, batch size 256, 5 seeds; null $=0.50$). Original models sit near the null, indicating no detectable memorisation prior to unlearning. NegGrad$+$ is the exception, increasing $\Mfour$ on the Residual MLP and FT-Transformer (bold; e.g.\ $0.60$ on Higgs with FT-Transformer), indicating displacement rather than removal.}
\label{tab:largescale}
\setlength{\tabcolsep}{5.6pt}
\begin{tabular}{lcccccc}
\toprule
 & \multicolumn{3}{c}{Covertype} & \multicolumn{3}{c}{Higgs} \\
\cmidrule(lr){2-4}\cmidrule(lr){5-7}
Method & MLP & Res.\ MLP & FT-Trans. & MLP & Res.\ MLP & FT-Trans. \\
\midrule
Original        & $0.510$ & $0.506$ & $0.498$ & $0.501$ & $0.496$ & $0.506$ \\
Gradient Ascent & $0.508$ & $0.499$ & $0.500$ & $0.495$ & $0.509$ & $0.512$ \\
NegGrad$+$      & $0.510$ & $\mathbf{0.523}$ & $\mathbf{0.529}$ & $0.518$ & $\mathbf{0.561}$ & $\mathbf{0.602}$ \\
Fine-Tuning     & $0.510$ & $0.505$ & $0.500$ & $0.501$ & $0.496$ & $0.504$ \\
SCRUB           & $0.509$ & $0.506$ & $0.487$ & $0.502$ & $0.499$ & $0.517$ \\
\bottomrule
\end{tabular}
\end{table}

\subsection{$\Mfour$ on image data}
\label{app:image_m4}
Table~\ref{tab:image_m4} reports $\Mfour$ on CIFAR-10, SVHN, and CIFAR-100 (10{,}000 samples per dataset, $\ff=\SI{5}{\percent}$, 5 seeds) for two architectures: a three-layer CNN and ResNet-18.

\begin{table}[H]
\centering
\footnotesize
\caption{$\Mfour$ on image datasets (CIFAR-10, SVHN, CIFAR-100; 10{,}000 samples, $\ff=\SI{5}{\percent}$, 5 seeds; mean $\pm$ SD; null $=0.50$). Original models sit near the null, indicating no detectable memorisation prior to unlearning. On ResNet-18, NegGrad$+$ and SCRUB produce substantial over-displacement (bold; values far below $0.50$), whilst the three-layer CNN is much less affected.}
\label{tab:image_m4}
\setlength{\tabcolsep}{7pt}
\begin{tabular}{lccc}
\toprule
Method & CIFAR-10 & SVHN & CIFAR-100 \\
\midrule
\multicolumn{4}{l}{\textbf{Three-layer CNN}} \\
Original         & $0.5066 \pm 0.0211$  & $0.5130 \pm 0.0079$  & $0.4915 \pm 0.0077$ \\
Gradient Ascent  & $0.5027 \pm 0.0122$  & $0.5023 \pm 0.0128$  & $0.5058 \pm 0.0118$ \\
NegGrad$+$       & $0.5052 \pm 0.0478$  & $0.5287 \pm 0.0181$  & $0.5370 \pm 0.0066$ \\
Fine-Tuning      & $0.4908 \pm 0.0100$  & $0.5090 \pm 0.0078$  & $0.4664 \pm 0.0098$ \\
SCRUB            & $0.5066 \pm 0.0377$  & $0.5328 \pm 0.0211$  & $0.5861 \pm 0.0538$ \\
\addlinespace
\multicolumn{4}{l}{\textbf{ResNet-18}} \\
Original         & $0.5010 \pm 0.0221$  & $0.5050 \pm 0.0165$  & $0.4928 \pm 0.0173$ \\
Gradient Ascent  & $0.5044 \pm 0.0082$  & $0.5199 \pm 0.0107$  & $0.5227 \pm 0.0154$ \\
NegGrad$+$       & $\mathbf{0.1598 \pm 0.0543}$  & $\mathbf{0.0335 \pm 0.0230}$  & $\mathbf{0.1593 \pm 0.1716}$ \\
Fine-Tuning      & $0.4720 \pm 0.0132$  & $0.4919 \pm 0.0143$  & $0.4398 \pm 0.0203$ \\
SCRUB            & $\mathbf{0.3689 \pm 0.1151}$  & $\mathbf{0.2044 \pm 0.1290}$  & $\mathbf{0.1131 \pm 0.0768}$ \\
\bottomrule
\end{tabular}
\end{table}

\subsection{$\Mfour$ on LFW face-identity unlearning}
\label{app:lfw}
Table~\ref{tab:lfw_m4} reports $\Mfour$ on the LFW face-identity experiment (4{,}324 images, 158 identities, 5 seeds). Forget sets comprise all images belonging to selected identities. The pre-unlearning row gives the diagnostic value before any unlearning, all far above the null and reflecting strong identity-level memorisation. With $N=5$ seeds, the minimum achievable Wilcoxon $p$-value is $0.062$; no condition reaches significance and $p$-values are therefore omitted.

\begin{table}[H]
\centering
\footnotesize
\caption{$\Mfour$ on LFW face-identity unlearning ($N=5$ seeds; mean $\pm$ SD). Forget sets comprise all images belonging to selected identities. The pre-unlearning row shows the diagnostic value before any unlearning is applied. Bold indicates substantial deviation from null ($|\Mfour-0.50|>0.15$).}
\label{tab:lfw_m4}
\setlength{\tabcolsep}{7pt}
\begin{tabular}{lccc}
\toprule
Method & $\ff=\SI{1}{\percent}$ & $\ff=\SI{5}{\percent}$ & $\ff=\SI{10}{\percent}$ \\
\midrule
Pre-unlearning   & $\mathbf{0.941 \pm 0.070}$ & $\mathbf{0.773 \pm 0.046}$ & $\mathbf{0.728 \pm 0.026}$ \\
\addlinespace
Gradient Ascent  & $\mathbf{0.906 \pm 0.146}$ & $\mathbf{0.703 \pm 0.020}$ & $\mathbf{0.728 \pm 0.091}$ \\
NegGrad$+$       & $0.650 \pm 0.158$          & $\mathbf{0.661 \pm 0.115}$          & $\mathbf{0.747 \pm 0.187}$ \\
Fine-Tuning      & $\mathbf{0.868 \pm 0.090}$ & $\mathbf{0.730 \pm 0.045}$ & $\mathbf{0.745 \pm 0.048}$ \\
SCRUB            & $\mathbf{0.235 \pm 0.183}$ & $0.590 \pm 0.205$          & $0.649 \pm 0.198$ \\
\bottomrule
\end{tabular}
\end{table}

\begin{figure}[H]
  \centering
  \includegraphics[width=\linewidth]{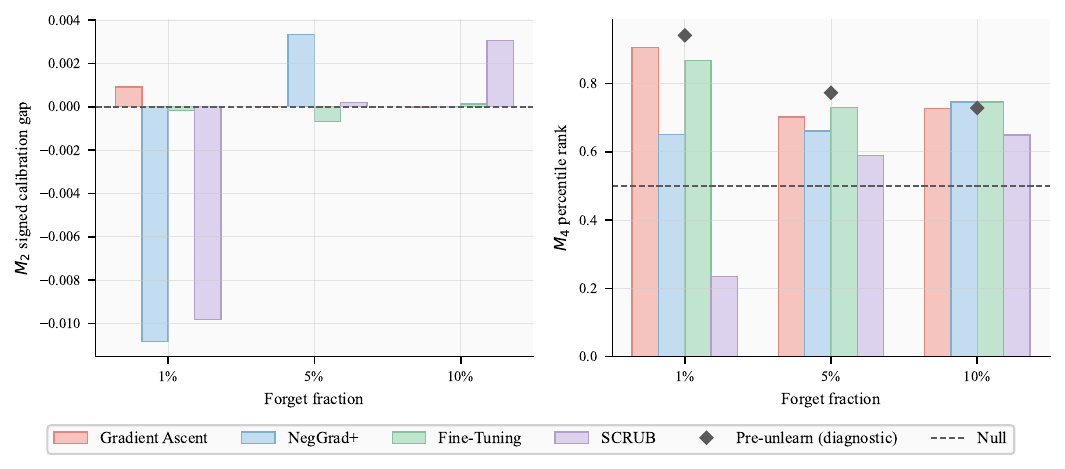}
  \caption{RULER on LFW face-identity unlearning (gender classification task, ResNet-18, $N=5$ seeds). Pre-unlearning $\Mfour$ (leftmost bar/row) is far above the $0.50$ null at every forget fraction, confirming the model encodes identity-level structure despite the gender-classification objective. After unlearning, Gradient Ascent and Fine-Tuning leave the identity signal largely intact; NegGrad$+$ partially reduces it; SCRUB substantially over-displaces at $\ff=\SI{1}{\percent}$. $\Mtwo$ values are close to zero with high variance; $\Mfour$ therefore carries the residual signal in this setting.}
  \label{fig:lfw}
\end{figure}

\subsection{Architectures for image and clinical-text diagnostics}
\label{app:arch_extra}
\begin{figure}[H]
  \centering
  \resizebox{\textwidth}{!}{%
  \begin{tikzpicture}[
      font=\tiny, every node/.style={align=center},
      img/.style={draw=rulermuted,line width=0.4pt,rounded corners=2pt,fill=white,minimum width=0.85cm,minimum height=0.55cm},
      blk/.style={draw=ruleraccent,line width=0.4pt,rounded corners=2pt,fill=ruleraccent!10,minimum width=0.9cm,minimum height=0.55cm},
      cblk/.style={draw=ruleraccent,line width=0.4pt,rounded corners=2pt,fill=ruleraccent!10,minimum width=2.1cm,minimum height=0.55cm},
      pen/.style={draw=forgetred,line width=0.9pt,rounded corners=2pt,fill=forgetred!12,minimum width=1.0cm,minimum height=0.55cm,font=\tiny\bfseries,text=forgetred},
      outbox/.style={draw=oraclegreen,line width=0.4pt,rounded corners=2pt,fill=oraclegreen!8,minimum width=0.9cm,minimum height=0.55cm},
      arr/.style={-{Stealth[length=1.6mm]},rulermuted,line width=0.35pt}]
    \node[font=\scriptsize\bfseries,text=ruleraccent,anchor=west] at (-7.3,1.0) {Three-layer CNN};
    \node[img] (in1) at (-6.6,0.4) {32$\times$32};
    \node[cblk, right=0.18cm of in1] (cb1) {Conv + ReLU + Pool};
    \node[blk, right=0.18cm of cb1] (f1) {FC$_1$};
    \node[pen, right=0.18cm of f1] (pen1) {FC$_2$\\\tiny 256-d};
    \node[outbox, right=0.18cm of pen1] (out1) {softmax};
    \foreach \a/\b in {in1/cb1, cb1/f1, f1/pen1, pen1/out1} \draw[arr] (\a) -- (\b);
    \node[font=\tiny\bfseries,text=forgetred] at (pen1.south) [below=0.02cm] {penultimate};
    \node[font=\scriptsize\bfseries,text=ruleraccent,anchor=west] at (-7.3,-1.0) {ResNet-18};
    \node[img] (in2) at (-6.6,-1.6) {image};
    \node[blk, right=0.15cm of in2] (stem) {Stem};
    \node[blk, right=0.12cm of stem] (s1) {S1};
    \node[blk, right=0.12cm of s1] (s2) {S2};
    \node[blk, right=0.12cm of s2] (s3) {S3};
    \node[blk, right=0.12cm of s3] (s4) {S4};
    \node[pen, right=0.15cm of s4] (gap) {GAP\\\tiny 512-d};
    \node[outbox, right=0.15cm of gap] (fc2) {FC\,+\,sm};
    \foreach \a/\b in {in2/stem, stem/s1, s1/s2, s2/s3, s3/s4, s4/gap, gap/fc2} \draw[arr] (\a) -- (\b);
    \node[font=\tiny\bfseries,text=forgetred] at (gap.south) [below=0.02cm] {penultimate};
  \end{tikzpicture}%
  }
  \caption{Image architectures used in the $\Mfour$ diagnostic. RULER extracts the activation just before the classifier: the second fully-connected layer for the three-layer CNN (256-d), and the post-GAP activation for ResNet-18 (512-d). The ResNet-18 backbone is also used for the face-identity experiment on LFW.}
  \label{fig:arch_image}
\end{figure}
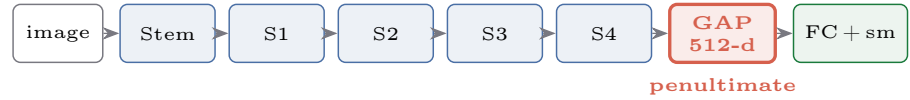

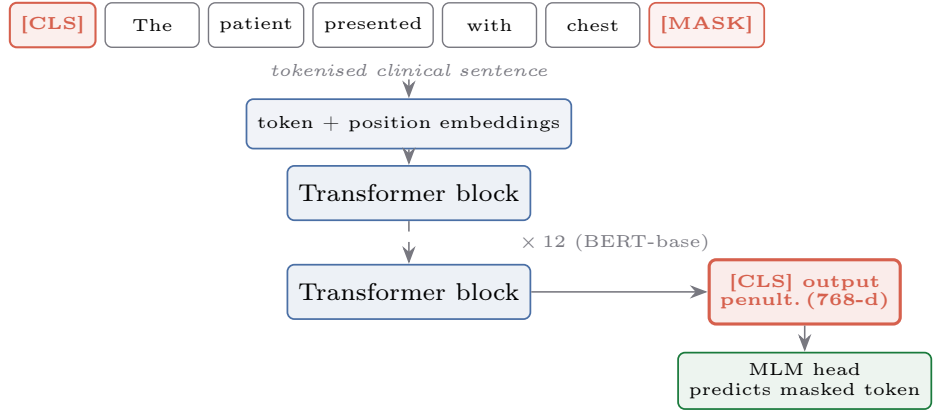
\begin{figure}[H]
  \centering
  \resizebox{\textwidth}{!}{%
  \begin{tikzpicture}[
      font=\scriptsize, every node/.style={align=center},
      tok/.style={draw=rulermuted,line width=0.4pt,rounded corners=2pt,fill=white,minimum width=0.95cm,minimum height=0.45cm,font=\tiny},
      cls/.style={draw=forgetred,line width=0.6pt,rounded corners=2pt,fill=forgetred!10,minimum width=0.85cm,minimum height=0.45cm,font=\tiny\bfseries,text=forgetred},
      emb/.style={draw=ruleraccent,line width=0.5pt,rounded corners=2pt,fill=ruleraccent!8,minimum width=1.9cm,minimum height=0.5cm,font=\tiny},
      tx/.style={draw=ruleraccent,line width=0.5pt,rounded corners=2pt,fill=ruleraccent!10,minimum width=1.4cm,minimum height=0.55cm,font=\scriptsize},
      pen/.style={draw=forgetred,line width=0.9pt,rounded corners=2pt,fill=forgetred!12,minimum width=1.55cm,minimum height=0.65cm,font=\tiny},
      mlm/.style={draw=oraclegreen,line width=0.5pt,rounded corners=2pt,fill=oraclegreen!8,minimum width=1.9cm,minimum height=0.5cm,font=\tiny},
      arr/.style={-{Stealth[length=1.8mm]},rulermuted,line width=0.45pt}]
    \node[font=\scriptsize\bfseries,text=ruleraccent,anchor=west] at (-6.1,2.05) {Bio\_ClinicalBERT};
    % input token row
    \node[cls] (cls) at (-5.6,1.4) {[CLS]};
    \node[tok, right=0.08cm of cls] (t1) {The};
    \node[tok, right=0.08cm of t1] (t2) {patient};
    \node[tok, right=0.08cm of t2] (t3) {presented};
    \node[tok, right=0.08cm of t3] (t4) {with};
    \node[tok, right=0.08cm of t4] (t5) {chest};
    \node[tok,fill=forgetred!8,draw=forgetred,line width=0.5pt,right=0.08cm of t5] (t6) {\textbf{\textcolor{forgetred}{[MASK]}}};
    \node[font=\tiny\itshape,text=rulermuted] at (-2.0,0.95) {tokenised clinical sentence};
    % encoder stack
    \node[emb] (emb) at (-2.0,0.4) {token + position embeddings};
    \node[tx] (tx1) at (-2.0,-0.3) {Transformer block};
    \node[font=\tiny,text=rulermuted,anchor=west] at (-1.0,-0.8) {$\times\,12$ (BERT-base)};
    \node[tx] (tx2) at (-2.0,-1.3) {Transformer block};
    % two branches from the final block: RULER extraction and MLM head
    \node[pen] (pen) at (2.0,-1.3) {\textbf{\textcolor{forgetred}{[CLS] output}}\\[-0.1em]\tiny\bfseries\color{forgetred}penult.\,(768-d)};
    \node[mlm] (mlm) at (2.0,-2.35) {MLM head\\[-0.05em]\tiny predicts masked token};
    % arrows
    \draw[arr] (-2.0,0.85) -- (emb.north);
    \draw[arr] (emb.south) -- (tx1.north);
    \draw[arr,dashed] (tx1.south) -- (tx2.north);
    \draw[arr] (tx2.east) -- (pen.west);
    \draw[arr] (tx2.south east) -- node[above,sloped,pos=0.45,font=\tiny,text=rulermuted]{[MASK] vector} (mlm.west);
  \end{tikzpicture}%
  }
  \caption{Bio\_ClinicalBERT. The training objective is masked language modelling: $15\%$ of the tokens in each input sentence are selected at random and the model is trained to recover the original word from the surrounding context, choosing from a vocabulary of around $30{,}000$ tokens. This is not classification: the model predicts words rather than assigning records to a fixed set of classes. RULER operates on the geometry of the [CLS] embedding regardless of the training objective.}
  \label{fig:arch_bert}
\end{figure}

Figs.~\ref{fig:arch_image} and~\ref{fig:arch_bert} show the image and Bio\_ClinicalBERT architectures used in Section~\ref{subsec:diagnostic}. RULER extracts the penultimate-layer activation just before the task-specific output head in every case.
\vspace{-0.5cm}

\vspace{-0.5cm}

\subsection{Reproducibility}
\label{app:repro}
The RULER library and tutorial are available at \url{https://github.com/gcosma/RULER}. All experiments use fixed random seeds propagated to dataset splitting (random state 999), forget-set sampling (random state 999), training initialisation (seeds 0 to 9), and unlearning (seed 100). For each training seed $i\in\{0,\ldots,9\}$, the original model $\modelO$ and retrain oracle $\modelR$ are trained with the same random seed. Same-seed oracle--original similarity has a grand mean of $0.993$ across ten datasets, compared with $0.431$ for differently seeded oracle pairs, confirming that shared initialisation produces similar representational geometry. Original models and retrain oracles are cached as \texttt{.pt} files after the initial training run, and all unlearning procedures operate from these identical cached weights.

\section*{Citing this work}
\begin{small}
\begin{verbatim}
@inproceedings{cosma2026ruler,
  title     = {{RULER}: Representation-Level Verification of Machine Unlearning},
  author    = {Cosma, Georgina and Finke, Axel},
  booktitle = {2nd Workshop on Machine Unlearning and Privacy Preservation
               ({WIPE-OUT}), {ECML} {PKDD}},
  address   = {Naples, Italy},
  year      = {2026},
  eprint    = {2605.27569},
  archivePrefix = {arXiv},
  note      = {To appear in Springer CCIS post-proceedings}
}
\end{verbatim}
\end{small}

\end{document}